# A transformer-BiGRU-based framework with data augmentation and confident learning for network intrusion detection

Jiale Zhang[1,2], Pengfei He[3], Fei Li[1,2], Kewei Li[1,2], Yan Wang[1,2], Lan Huang[1,2], Ruochi Zhang[1,4,*], Fengfeng Zhou[1,2,5,*].

1 Key Laboratory of Symbolic Computation and Knowledge Engineering of Ministry of Education, Jilin University, Changchun, Jilin, China, 130012.

2 College of computer Science and Technology, Jilin University, Changchun, Jilin, China, 130012.

3 School of Artificial Intelligence, Xidian University, Xi'an, Shaanxi, China,710126

4 School of Artificial Intelligence, Jilin University, Changchun, Jilin, China, 130012.

5 School of Biology and Engineering, Guizhou Medical University, Guiyang 550025, Guizhou, China.

* Correspondence may be addressed to Fengfeng Zhou: FengfengZhou@gmail.com or ffzhou@jlu.edu.cn. Correspondence may also be addressed to Ruochi Zhang: zrc720@gmail.com.

# Abstract

In today's fast-paced digital communication, the surge in network traffic data and frequency demands robust and precise network intrusion solutions. Conventional machine learning methods struggle to grapple with complex patterns within the vast network intrusion datasets, which suffer from data scarcity and class imbalance. As a result, we have integrated machine learning and deep learning techniques within the network intrusion detection system to bridge this gap. This study has developed TrailGate, a novel framework that combines machine learning and deep learning techniques. By integrating Transformer and Bidirectional Gated Recurrent Unit (BiGRU) architectures with advanced feature selection strategies and supplemented by data augmentation techniques, TrailGate can identifies common attack types and excels at detecting and mitigating emerging threats. This algorithmic fusion excels at detecting common and well-understood attack types and has the unique ability to swiftly identify and neutralize emerging threats that stem from existing paradigms.

# 1 Introduction

The rapid advancement of information and communication technologies, coupled with the proliferation of the Internet of Things (IoT), has resulted in the integration of computers and networks into various societal aspects [1]. This integration has led to a significant increase in network traffic and has attracted numerous attackers [2]. However, the development of digital communication technologies and the widespread use of IoT have significantly increased the volume and complexity of network traffic. Consequently, this complexity has resulted in a corresponding rise in network intrusions, presenting significant challenges to cybersecurity. New attacks have been devised to breach security measures for financial gains [3]. At the same time, traditional network intrusion detection (NID) methods, which often rely on signature-based detection and expert-defined features, are increasingly inadequate in identifying sophisticated and novel attacks. The 2023 National Internet Emergency Center (CNCERT/CC) reports have highlighted new malicious attacks targeting the mobile internet and increasing security vulnerabilities [4]. This trend underscores the growing complexity of the network environment and the intensification of network intrusions. Therefore, it is imperative to develop innovative prevention mechanisms and intrusion detection to counter these evolving threats [5].

Traditional NID methods typically use signature-based schemes or machine learning algorithms. Signature-based schemes rely on pre-defined intrusion-specific patterns to detect network intrusion events [6, 7]. However, new intrusion types may go undetected because their signatures are unavailable in the database. The NID problem can be defined as a binary classification task between normal interaction accesses and anomalies [8] or a multi-class classification task of different intrusion types [9]. Feature selection algorithms may be utilized to improve the expert-defined features for better classification performances [10]. Machine learning NID methods depend on expert-defined features [5] and often experience high false positive rates [11]. The main challenge in NID is detecting new and previously unseen attack patterns, which traditional methods usually fail to recognize due to their reliance on pre-defined signatures and features. This challenge is intensified by the dynamic nature of network environments and the rapid evolution of attack strategies.

Deep learning algorithms are compelling, allowing researchers to uncover hidden features within complex datasets and gaining broad interest in the NID [12]. Studies have explored the NID task using various deep learning algorithms, such as autoencoder [13], restricted Boltzmann machine [14], and recurrent neural network [15]. However, these deep learning-based NID methods often encounter challenges, including limited labeled data, high training costs, and insufficient generalization when facing novel or complex intrusion patterns.

Meanwhile, Transformer-based architectures have achieved remarkable success across a broad spectrum of domains, including natural language processing, computer vision, and even specialized applications like underwater image captioning [16]. Their ability to model global dependencies and capture contextual relationships has made them a powerful tool for many tasks. However, most existing NID methods have yet to fully leverage the Transformer's capacity for long-range dependency modeling and attention mechanisms, which limits their ability to detect complex and subtle intrusion patterns. Similarly, Large model have emerged as effective solutions in areas such as discriminative image enhancement and domain adaptation [17]. Inspired by these developments, researchers have begun adopting Transformer and large model techniques for cybersecurity tasks, including network intrusion detection, to better model temporal sequences and uncover subtle anomaly patterns. These advances demonstrate the scalability and cross-domain potential of such models, further justifying their application in complex network environments.

To overcome the limitations of existing methods, this study, for the first time, integrates confident learning into the feature selection process of network intrusion detection (NID), aiming to identify mislabeled or anomalous samples in the dataset and effectively select features with stronger generalization capabilities. In addition, a universal feature selection framework suitable for heterogeneous datasets is designed, further enhancing the applicability and effectiveness of the detection system.

This study aims to develop a robust NID framework that can detect a wide range of attack types, including those that have not been previously encountered. The proposed method combines deep learning techniques with feature selection and data augmentation strategies to enhance detection accuracy, especially for rare and sophisticated attack types such as User-to-Root (U2R) and Remote-to-Local (R2L) intrusions.

This study provides the following contributions:

1) First, we have developed a new two-stage framework, TrailGate, which integrates machine learning and deep learning techniques for NID tasks. This combination leverages both technology's strengths to enhance the accuracy of detecting common and rare intrusion types.

2) Second, we have incorporated a robust feature selection method that can be used as a universal feature selection framework for heterogeneous datasets, including numerical and non-numerical data. This method combines adaptive data augmentation techniques to improve model training efficiency and ensure effectiveness when tested on independent and imbalanced datasets, addressing the common class imbalance problem in NID tasks.

3) In addition, the proposed TrailGate framework has been rigorously evaluated through binary and multi-class classification tasks, with results showing superior performance compared with existing research. Notably, TrailGate has made significant progress in detecting U2R and R2L attack types, which are traditionally difficult to identify. These enhanced features fill a critical gap in current intrusion detection systems, making TrailGate particularly effective in real-world network security applications.

4) For the first time, confident learning has been integrated into the data preprocessing pipeline to select samples with labeling errors or noise for feature selection. These strategies filter out more robust features, reducing false alarm rates and improving detection accuracy to enhance the model's robustness, especially in real-world network environments where data quality may vary.

## 2 Related work

The realm of NID has been extensively investigated. The NID studies mainly focus on binary classification tasks, distinguishing between normal and abnormal network traffic. Alazab et al. introduced the Moth-Flame Optimizer algorithm to fine-tune Support Vector Machine (SVM) parameters, significantly improving intrusion detection in network traffic [18]. The innovation of their approach lies in the adaptive optimization of SVM's hyperparameters, leading to a substantial improvement in accuracy. Their experiments achieved an accuracy of 89.7% on the KDDTest+ dataset and demonstrated reduced false positives compared with traditional SVM-based approaches. However, the method had limitations in detecting minority classes like U2R and R2L attacks. Mushtaq et al. combined an autoencoder with a long short-term memory (LSTM) network to develop a two-stage intrusion detection model [19]. The autoencoder was utilized for unsupervised feature extraction, while the LSTM was used for temporal sequence learning. Their approach achieved an impressive accuracy of 87.37% on the KDDTest+ dataset, with improved performance in detecting Denial of Service (DoS) attacks. However, it struggled with more complex attack types like U2R due to relying on temporal correlations alone. The Dual Intrusion Detection System (Dual-IDS) approach employed a machine learning-based strategy and a bagging-based gradient boosting decision tree model to improve intrusion detection accuracy [20]. The key innovation was using an ensemble method to address overfitting and improve the detection of imbalanced datasets. Dual-IDS achieved 91.57% accuracy on the KDDTest+ dataset and 82.35% on the KDDTest-21 dataset, outperforming single-model approaches. However, its computational complexity made it less suitable for real-time applications. Emil Selva et al. proposed a deep learning-based feature fusion method for detecting distributed denial of service attacks in cloud computing [21]. The method combines multiple feature extraction techniques to build a more powerful representation of network traffic for Deep Neural Network (DNN) to improve the detection of distributed denial of service attacks. Experimental results showed 93.04% accuracy on the KDDTest+ dataset but reduced performance on other intrusions, indicating a lack of generalization. The semi-supervised MANomaly system uses Mutual Adversarial Networks (MANs) to improve anomaly detection. This system employs dual MANs for generating synthetic data and discrimination [22]. Using adversarial networks allows for better handling of unlabeled data, making this system suitable for real-world applications where labeled data is limited. MANomaly achieved 92.75% accuracy on the KDDTest+ dataset and 88.11% accuracy on the KDDTest-21 dataset, with a strong performance in detecting new and unknown attack types. However, the reliance on synthetic data generation introduced potential instability in adversarial training.

The following studies have extended beyond binary classification to focus on detecting multiple intrusion types. The Dynamic Ensemble Approach (DUEN) addressed the challenge of class imbalance in NID using a dynamic weighting scheme to adjust the influence of individual classifiers based on their performance on minority classes [23]. DUEN's breakthrough innovation lies in its dynamic adjustment mechanism, which has significantly improved the detection of underrepresented attack types like U2R and R2L. Achieving an impressive overall accuracy of 82.6% on the NSL-KDD dataset, this model outperformed static ensemble methods by a wide margin. By contrast, the ROULETTE model introduced a cutting-edge neural attention multi-output framework to classify network traffic into multiple classes, including normal accesses, known attacks, and unknown attacks [24]. Its attention mechanism allowed the system to concentrate on the most critical parts of the input data for more precise multi-class classification, resulting in an accuracy of 81.5% on the NSL-KDD dataset, with a solid performance in distinguishing between known and unknown attack types, although its effectiveness diminished when confronted with highly imbalanced data. In addition, the Modified Density Peak Clustering with Deep Belief Networks (MDPC-DBN) model combined unsupervised clustering with deep belief networks to improve the multi-class classification of network traffic [25]. With its innovative integration of density peak clustering for feature clustering, this model helped to separate different attack types before classification and achieved 82.08% accuracy on the KDDTest+ dataset and 66.18% accuracy on the KDDTest-21 dataset, particularly effective in detecting DoS. While these models have demonstrated remarkable accuracy and efficiency, it is essential to consider their suitability for real-time applications, as their performance decreased with high-dimensional data. The SAAE-DNN model combined a stacked autoencoder with a deep neural network to improve feature extraction for multi-class intrusion detection tasks [26]. Its novelty lies in its unsupervised pre-training, which enhances the deep neural network's ability to classify multiple intrusion tasks. The model achieved 82.1% accuracy on the KDDTest+ dataset and demonstrated robust performance across attack types such as DoS and Probe. However, its performance decreased when dealing with rare attack types like U2R, highlighting the challenge of handling imbalanced datasets.

In addition to the above methods, several recent studies have explored AI-driven and cross-domain strategies to further enhance network intrusion detection, especially under emerging network paradigms such as IoT and 5G. For instance, Sharma et al. conducted a comparative study of AI-based prediction models for network security[27], highlighting the effectiveness of intelligent classifiers in adapting to evolving attack patterns. Similarly, a cognitive security framework was proposed for detecting intrusions in IoT and 5G environments using deep learning techniques [28], showing strong potential in handling heterogeneous and large-scale network traffic with improved generalization. From an optimization perspective, Dalal et al. [29] introduced an optimized LightGBM model to address security and privacy issues in cyber-physical systems, which are becoming increasingly important in real-time applications. Furthermore, to improve energy efficiency in mobile ad hoc networks, Hooda et al. [30] proposed a power-aware routing approach based on cluster-head selection using a gateway table. Although these works are not directly focused on traditional NID datasets like KDD or NSL-KDD, they offer important insights and demonstrate the broader applicability of AI and deep learning models in various cybersecurity contexts.

Collectively, these emerging approaches emphasize the importance of model optimization, energy efficiency, and adaptability to complex environments, which are essential considerations for future NID systems. Inspired by these advancements, our work integrates multiple learning paradigms and optimization strategies to improve detection performance across diverse and imbalanced intrusion scenarios.

The above-reviewed studies incorporate different strategies to address class-imbalanced NID datasets. These strategies include feature selection, data augmentation, and the development of innovative classifiers. We agree with the effectiveness of these methods and plan to harness the strengths of machine learning and deep learning for the NID task.

Despite significant advancements in various NID techniques, technical gaps remain in the existing approaches. Conventional methods relied heavily on pre-defined features and signatures, making them less effective in detecting novel or unknown attack types. While deep learning methods have shown promise in feature extraction and classification tasks, they still struggle with issues such as data imbalance and high training costs. Many current techniques exhibit lower accuracy in detecting complex attack types, such as U2R and R2L, which are critical in modern, constantly evolving network environments.

The identified technical gaps led to the design of the proposed two-stage NID framework. Our approach integrates machine learning and deep learning techniques with our proposed feature selection method and data augmentation strategy to enhance detection accuracy, especially for rare and complex attack types. By addressing the limitations of current approaches, our framework significantly improves the detection of various attacks while decreasing false positive rates, ultimately advancing overall network security capabilities.

## 3 Materials and methods

### 3.1 Dataset

The NSL-KDD dataset [31] was a widely recognized benchmark in network intrusion detection. This dataset is an improved version of the KDDCup99 dataset and addresses several limitations, such as removing duplicate and redundant records [32]. While it may not perfectly reflect real-world traffic data, the benchmark dataset represents our purposes. This study utilizes the KDDTrain+ subset as the training set, with both KDDTest+ and KDDTest-21 subsets as the independent test sets. Table 1 summarizes the details of data within these datasets.

This study covers binary and multi-class classification tasks using independent test sets. The NSL-KDD dataset contains 41 features, consisting of 3 discrete and 38 continuous-valued features, as detailed in Table 2. It encompasses various types of network traffic, including regular traffic and four main types of network intrusions, each further divided into multiple sub-categories, as shown in Table 3. Notably, the training set includes 22 sub-types of attacks, while the independent test sets encompass 37 sub-types. This difference presents a significant challenge for detecting the sub-types of attacks in the test sets, which are not present in the training set. This situation also creates an intriguing scenario for evaluating our proposed model.

Table 1. Summary of the NSL-KDD benchmark dataset. The numbers in the grids indicate the numbers of samples, outlining the three subsets of the NSL-KDD dataset.

| Dataset | Normal | Network Intrusion Type | | | | Total |
|---|---|---|---|---|---|---|
| | | DoS | Probe | U2R | R2L | |
| KDDTrain+ | 67343 | 45927 | 11656 | 52 | 995 | 125973 |
| KDDTest+ | 9711 | 7458 | 2421 | 200 | 2754 | 22544 |
| KDDTest-21 | 2152 | 4342 | 2402 | 200 | 2754 | 11850 |

Table 2. Types of feature values in the dataset. "D" and "C" represent the discrete and continuous-valued features, respectively.

| Feature name | Type | Feature name | Type |
|---|---|---|---|
| duration | C | is_guest_login | C |
| protocol_type | D | count | C |
| service | D | srv_count | C |
| flag | D | serror_rate | C |
| src_bytes | C | srv_serror_rate | C |
| dst_bytes | C | rerror_rate | C |
| land | C | srv_rerror_rate | C |
| wrong_fragmennt | C | same_srv_rate | C |
| urgent | C | diff_srv_rate | C |
| hot | C | srv_diff_host_rate | C |
| num_failed_logins | C | dst_host_count | C |
| logged_in | C | dst_host_srv_count | C |
| num_compromised | C | dst_host_same_srv_rate | C |
| root_shell | C | dst_host_diff_srv_rate | C |
| su_attempted | C | dst_host_same_src_port_rate | C |
| num_root | C | dst_host_srv_diff_host_rate | C |
| num_file_creations | C | dst_host_serror_rate | C |
| num_shells | C | dst_host_srv_serror_rate | C |
| num_access_files | C | dst_host_rerror_rate | C |
| num_outbound_cmds | C | dst_host_srv_rerror_rate | C |
| is_host_login | C | | |

Table 3. The types and sub-types of attacks in the training (KDDTrain+) and test (KDDTest+ and KDDTest-21) datasets.

| Type | Sub-type | Types of attack | |
|---|---|---|---|
| | | KDDTrain+ | KDDTest+ and KDDTest-21 |
| Normal | Normal | Normal | Normal |
| Attack | DoS | pod, land, teardrop, neptune, back, smurf | back, pod, land, smurf, teardrop, mailbomb, neptune, processtable, udpstorm, apache2, worm |
| | Probe | portsweep, ipsweep, satan, nmap | saint, portsweep, ipsweep, satan, nmap, mscan, |
| | R2L | multihop, ftp_write, warezmaster, phf, guess_passwd, spy, imap, warezclient, | ftp_write, guess_passwd, imap, multihop, phf, warezmaster, xlock, xsnoop, snmpguess, snmpgetattack, sendmail, named，httptunnel |
| | U2R | buffer_overflow, rootkit, loadmodule, perl | buffer_overflow, rootkit, loadmodule, perl, sqlattack, xterm, ps |

The dataset is preprocessed before being loaded to the downstream analyzing modules. The study first focuses on classifying types and sub-types, as shown in Table 3. Moreover, non-numerical features such as protocol_type, flag, and service are converted into a list of ordered numerical integers (0, 1, 2, …, $k$-1), where $k$ is the number of this feature's unique values. Finally, the min-max scaling is applied to scale each continuously-valued feature into the range [0, 1] using $X' = (X - X_{min})/(X_{max} - X_{min})$, where $X'$ and $X$ are the scaled and original value of a feature, respectively, and $X_{max}$ and $X_{min}$ are the maximum and minimum values of this feature, respectively.

## 3.2 Data augmentation

We used the Adaptive Synthetic Sampling Approach for Imbalanced Learning (ADASYN) method [33] to address the challenge of significant sample imbalance in our training dataset for the classification tasks. This approach is essential because it mitigates the model's inclination toward the majority class of samples during the learning process. Figure 1 shows the effectiveness of ADASYN in altering the sample distribution of our training set. In particular, TrailGate adopts a strategy of oversampling all classes except the majority class, ensuring that minority categories are sufficiently represented during training.

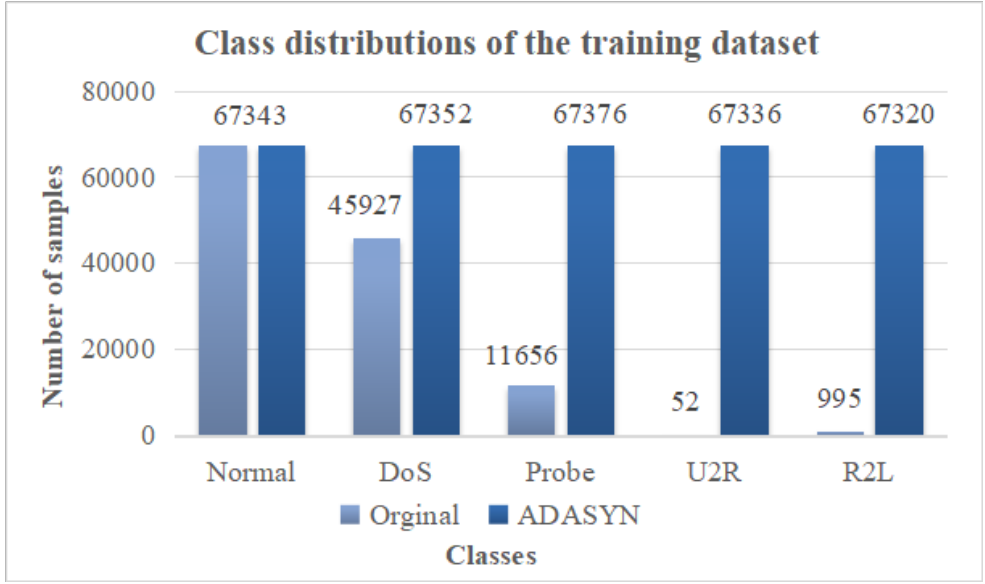

Figure 1. Sample distributions of the training set before and after sample augmentation using ADASYN. The horizontal axis displays the classes of samples, and the vertical axis represents the number of samples in each class.

ADASYN is particularly good at assigning different weights to different minority sample classes and automatically determines the required quantity of synthetic samples for each minority class. This functionality represents an improvement over the Synthetic Minority Over-sampling Technique (SMOTE) algorithm [34] because ADASYN focuses on the minority samples located at the decision boundary and enhances the overall class distribution by creating new samples.

## 3.3 Information gain

The dataset is a fascinating mix of discrete and continuous-valued features. We used robust metrics of calculation gain [35, 36] to uncover the most impactful feature. Information gain is determined by measuring the reduction in entropy resulting from partitioning the dataset using a particular feature. Essentially, features with higher information gain significantly impact the dataset, leading to a more distinct and less uncertain data division.

While information gain traditionally applies to discrete data, we have undertaken a dual-phase analysis. For discrete features, the information gain is calculated directly. However, we first convert the continuous-valued features into discrete categories through discretization. For continuous values, we sort the values of each feature in ascending order and select a splitting node, which is the midpoint of any two adjacent values. After obtaining the information entropy and information gain of all splitting points, we choose the node with the highest information gain as the splitting node. Then, the information gained for the continuous features is calculated on the discretized values, and the pseudo-code is detailed in Table 4.

Then, we ranked features based on the descendent order of their information gain values. This approach allows us to identify the most influential features for classification and gain valuable insights from the dataset.

Table 4. Pseudo-code for calculating the information gain.

---

**Algorithm. Calculation of Information Gain.**

**Input:** The dataset KDDTrain+ $Tr=\{X, Y\}$, where $X \in R^d$ is a $d$-dimensional vector of features; $N$ is the number of samples

**Output:** Information gain for each feature $Info(X)$

1: Divide the $Tr$ into only containing discrete features datasets $Dis\_Tr = \{dis\_X, Y\}$ and only containing continuous features datasets $Con\_Tr = \{con\_X, Y\}$;

  $dis\_X\_list=\{K_1, K_2, K_3\}$ represents the set of discrete feature in $Dis\_Tr$

  $con\_X\_list=\{K_1, K_2, K_3,…,K_{38}\}$ represents the set of continuous feature in $Con\_Tr$

  $n$ represents the number of categories for each feature

2: **function: IG_dis**

  **for** y in unique_value(Y) **do**

  $$H(Y) = -\sum_{x_i=1}^{n}[P(y) * \log P(y)]$$

  $P(y)$ represents the probability that the target variable $Y$ takes the value $y$

  **end for**

  **for** $X$ in $dis\_X\_list$ **do**

   **for** $x_i$ in unique_value(X) **do**

   $$H(Y|X = x_i) = -\sum_{x_i=1}^{n}[P(y|X = x_i) * \log P(y|X = x_i)]$$

   $P(y|X = x_i)$ represents the probability that the target variable $Y$ takes the value $y$ under the condition that the feature $X$ takes the value $x_i$

   where the unique_value(X) represents the category value set of discrete feature $X$

   **end for**

  Calculate the information gain for each feature IG_dis(X)

  $IG\_dis(X) = H(X) - \sum_{x_i=1}^{n}[P(x) * H(Y|X = x_i)]$

  **end for**

return IG_dis(*X*)
3：function: IG_con
　　for *y* in unique_value(*Y*) do
　　　　$H(Y) \mathrel{+}= -\sum_{x_i=1}^{n}[P(y) * \log P(y)]$
　　　　$P(y)$ represents the probability that the target variable *Y* takes the value *y*
　　end for
　　for *i* <- 1 to len(*con_X_list*) do
　　　　feature <- *con_X_list*[*i*]
　　　　sort_feature <- **sort**(*feature*)
　　　　threshold <- (sort_feature[*inde*-1]+sort_feature[*inde*])/2 **for** *inde* <-1 to len(*feature*)
　　　　thre_set <- remove_duplicates(*threshold*)
　　　　if max_value(*feature*) in *thre_set*
　　　　　　remove_element(*thre_set*, max_value(*feature*))
　　　　if min_value(*feature*) in *thre_set*
　　　　　　remove_element(*thre_set*, min_value(*feature*))
　　　　IG_con(*X*) = 0
　　　　for *thre* in *thre_set* do
　　　　　　lower = get_feature_values_below_threshold(*feature*, *thre*)
　　　　　　higher = get_feature_values_above_threshold(*feature*, *thre*)
　　　　　　for l in unique_value(*lower*) do
　　　　　　　　$H(lower) = -\sum_{l=1}^{n}[P(y|X=l) * \log P(y|X=l)]$
　　　　　　end for
　　　　　　for h in unique_value(*higher*) do
　　　　　　　　$H(higher) = -\sum_{h=1}^{n}[P(y|X=h) * \log P(y|X=h)]$
　　　　　　end for
　　　　　　condi_H = (len(lower)/*N*)*H(lower)+ (len(higher)/*N*)*H(higher)
　　　　　　IG(*X*) = H(*Y*)-*condi_H*
　　　　　　IG_con(*X*) <- **max**(*pre_H*, IG(*X*))
　　　　end for
　　end for
　　return IG_con(*X*)

8：**Output**：Info(*X*) <- **sort**(IG_dis(*X*)+IG_con(*X*))

## 3.4 Inter-feature correlations

We analyze the correlations between the features in the training data by calculating the Pearson correlation coefficient (PCC) for each pair of features, calculated for each pair of features. The PCC value range of [-1, 1], with values closer to 1, indicates a strong correlation between the two corresponding features. A negative value suggests an inverse correlation, whereas a positive value indicates a direct correlation.

However, a high correlation between two features can lead to redundancy in information, which may impact the predictive efficacy of the model. We use the metric PCC to detect inter-feature correlations, ultimately reducing the feature dimensionality [37].

## 3.5 Confident learning

Confident learning, proposed by Northcutt et al. at Google, aims to identify mislabeled samples within a training set [38]. This approach estimates the joint distribution between assigned (known) and true (unknown) labels without requiring hyperparameter tuning. It uses cross-validation to determine the predictive probabilities of samples and has the distinct advantage of being model-agnostic.

This study utilizes confidence learning to eliminate noise samples in the subsequent feature selection process and to choose samples that may be mislabeled or abnormal in the data-augmented training set. This technique improves the robustness of the feature selection process by introducing noise data, which is particularly important when dealing with imbalanced datasets and inconsistent inter-class data distributions. The confidence learning algorithm identifies mislabeled data points and adjusts their impact during training, ensuring that the selected features are not biased toward outliers or noise. Consequently, the model can generalize efficiently to unseen data and reduce the possibility of overfitting. By employing confidence learning, TrailGate can select features to improve prediction accuracy and shorten training time by reducing the overall feature set. The workflow of confidence learning is shown in Figure 2.

The process starts with obtaining the predicted probability $\tilde{P}$ for each sample during cross-validation. For the original label $y_t$ and a sample $x$, we compute the average probability for each category as the confidence threshold $A_j$, as outlined in the following equation. In this process, $X$ indicates the sample space, and $\theta$ denotes the parameters of the predictive model.

$$A_j = \frac{\sum_{x \in X_{y_t=j}} \tilde{P}(y_t=j;x;\theta)}{|X_{y_t=j}|} \quad (1)$$

We determine each sample's predicted class label using the following equation. The class label with the highest prediction probability that exceeds its confidence threshold is selected. In this context, $y_p$ represents the predicted label.

$$\tilde{X}_{y_t=i,y_p=j} := \{x \in X_{y_t=i}: \tilde{P}(y_t=j;x;\theta) \geq A_j, j = argmax(\tilde{P}(y_t=l;x;\theta))\} \quad (2)$$

A counting matrix between the predicted and real class labels is formed:

$$C_{y_t,y_p}[i][j] := |\tilde{X}_{y_t=i,y_p=j}| \quad (3)$$

Normalization of this counting matrix normalizes the joint distribution of predicted and real class labels, where $m$ represents the number of distinct class labels.

$$D_{y_t=i,y_p=j} = \frac{\frac{C_{y_t=i,y_p=j} \cdot |X_{y_t=i}|}{\sum_{j \in [m]} C_{y_t=i,y_p=j}}}{\sum_{i \in [m], j \in [m]}(\frac{C_{y_t=i,y_p=j} \cdot |X_{y_t=i}|}{\sum_{k \in [m]} C_{y_t=i,y_p=k}})} \quad (4)$$

Finally, we filter out data with discrepancies between predicted and real class labels, considering them abnormal. The goal is to refine the training set to include the most reliable samples and ensure optimal performance on the independent test sets.

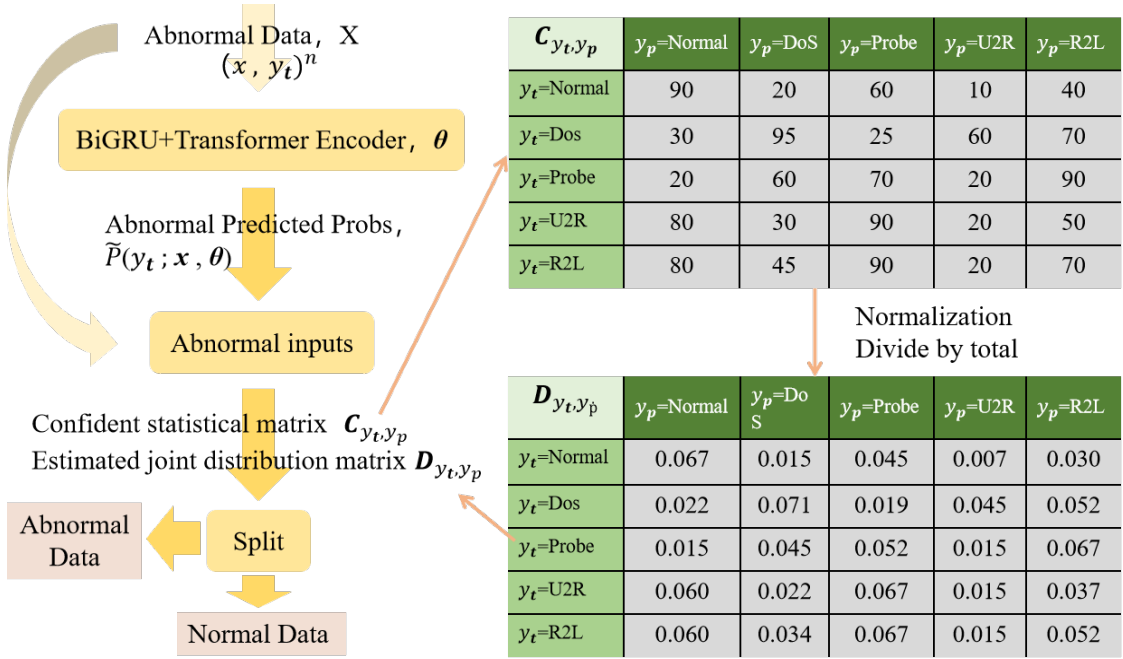

Figure 2. Confident Learning Flowchart.

## 3.6 BiGRU+Transformer

We have developed a deep learning framework that combines a bi-directional gated recurrent unit (BiGRU) with an encoder block from the Transformer architecture [39]. Figure 3 illustrates the structural framework. The BiGRU processes the input data forward and backward, capturing both short-term and long-term dependencies.

The BiGRU's bidirectional nature allows the model better to understand the temporal relationship between network traffic events, making it particularly suitable for detecting complex attack patterns such as U2R and R2L. In addition, the Transformer Encoder further enhances the BiGRU's output by focusing on the most critical features of the data using a multi-head attention mechanism. This attention mechanism dynamically assigns weights to different parts of the input sequence, allowing the model to prioritize critical features and ignore irrelevant noise. This framework improves the model's performance on complex multi-class classification tasks, such as distinguishing various intrusions.

In the independent test sets, our model is designed to handle various unseen sub-type attacks and improve predictive capabilities for future scenarios. We have integrated a lightweight BiGRU with the Transformer encoder architecture to enhance the model's sensitivity to anomalous data. The attention network and fully connected layers are harnessed to refine the extracted features for optimized performance. The experimental data effectively supports our model in mitigating classification challenges associated with complex sample distributions.

Suppose we consider an input sequence $X = \{X_1, X_2, \ldots, X_t\}$ and the corresponding output sequence of BiGRU $y = \{y_1, y_2, \ldots, y_t\}$, the following equations govern the dynamics of the input-output relationship at time step *t* and *t-1*.

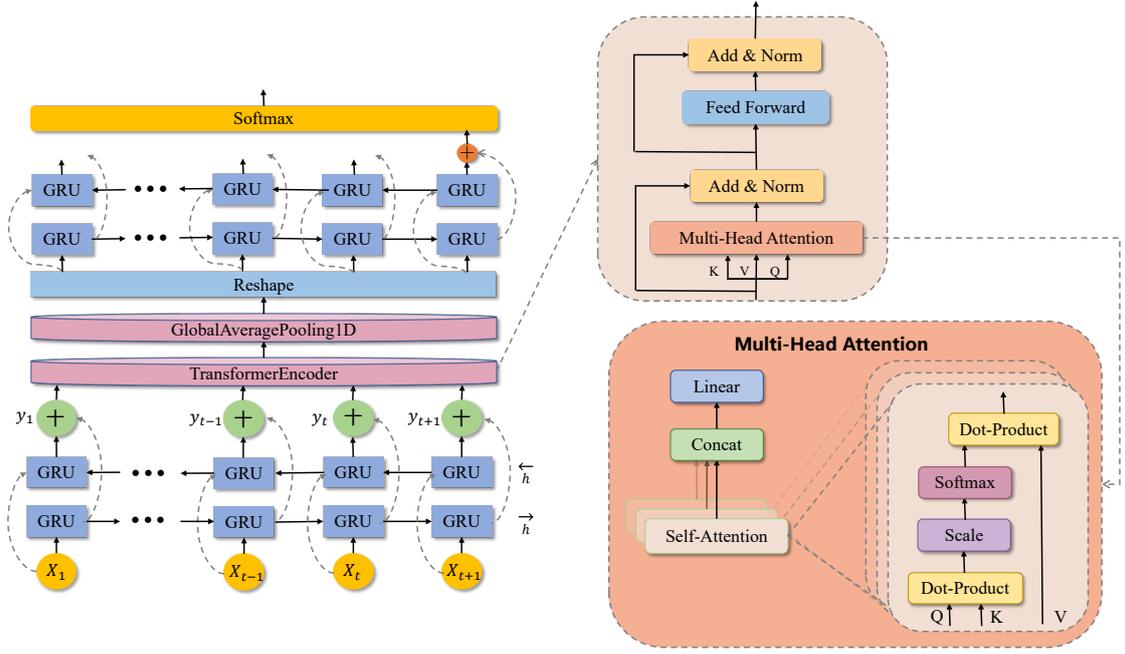

Figure 3. Framework diagram of the BiGRU+Transformer model. The Transformer's encoder module and its multi-head attention are illustrated with further details.

The hidden state $\overrightarrow{h_{t-1}}$ signifies the previous moment in the forward direction and the hidden state. Moreover, $\overleftarrow{h_{t+1}}$ represents the previous moment in the reverse direction. The current hidden states in the forward and reverse directions are calculated as follows:

$$\overrightarrow{h_t} = H(W_{\overrightarrow{h}} \cdot [\overrightarrow{h_{t-1}}, X_t] + b_{\overrightarrow{h}}) \quad (5)$$

$$\overleftarrow{h_t} = H(W_{\overleftarrow{h}} \cdot [\overleftarrow{h_{t+1}}, X_t] + b_{\overleftarrow{h}}) \quad (6)$$

The output $y_t$ at the current time point is formulated as follows:

$$y_t = W_y \cdot [\overrightarrow{h_t}, \overleftarrow{h_t}] + b_y \quad (7)$$

Figure 3 shows the Transformer's encoder block. Ten-fold cross-validation is employed to train this network using the cross entropy as the loss function.

## 3.7 Overall framework

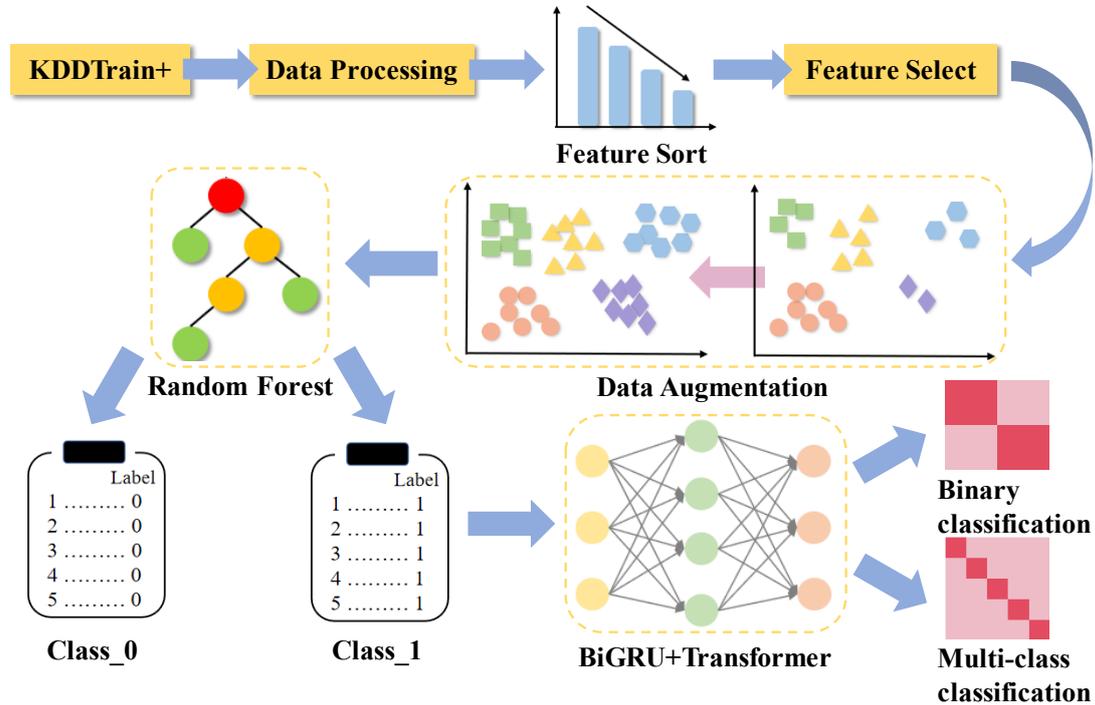

Figure 4. The framework of the proposed TrailGate model.

Table 5. Variable definitions are involved in the pseudo-code of TrailGate.

| Variable | Definition |
| --- | --- |
| $n$ | The number of training samples |
| $X\{X_1, X_2 \ldots X_{41}\}$ | Characteristics of the original training set |
| $Y\{Y_1, Y_2 \ldots Y_n\}$ | Labels of the original training set |
| $X_p\{X_{p1}, X_{p2} \ldots X_{p41}\}$ | Features preprocessed by data, such as Feature Encoding、Min-Max scale |
| $Y_p\{Y_{p1}, Y_{p2} \ldots Y_{pn}\}$ | Labels preprocessed by label conversion |
| $[X_e, Y_e]$ | Processed data with Data augmentation |
| $[X_s, Y_s]$ | Data filtered by information gain sorting and correlation coefficient |
| $[Ab_x, Ab_y]$ | Abnormal data selected through confident learning |
| $[X_k^a, Y_k^a]$ | Cross validation partitioned training set |
| $[X_k^t, Y_k^t]$ | Cross validation partitioned validation set |
| $[X_{fs}, Y_{fs}]$ | Features after feature selection |
| $[X_f, Y_f]$ | Samples and features after feature selection and data augmentation in Training set |
| $[X'_f, Y'_f]$ | Samples and features after feature selection in test set |

Table 6. Pseudo-code of the proposed TrailGate framework.

**Algorithm. The TrailGate framework.**

**Input:** The datasets KDDTrain+ $Tr=\{X, Y\}$, and KDDTest+ $Te=\{X', Y'\}$, where $X \in R^d$ and $X' \in R^d$ represent $d$-dimensional vector of features; a prediction model $Q$.
**Output:** Prediction confusion matrix of $Q$.

1: **function** $X_p \leftarrow$ FeatureEncoding($X$['protocol_type','service','flag'])、Min-Max($X$)
2: **function** $Y_p \leftarrow$ LabelEncoding($Y$)
3: **function** $[X_e, Y_e] \leftarrow$ ADASYN($X_p, Y_p$)
4: **function** $[X_s, Y_s] \leftarrow$ Information Gain($X_p, Y_p$)、Correlation Matrix($X_p, Y_p$)
5: **function** $[Ab_x, Ab_y] \leftarrow$ Confident Learning($X_p, Y_p$);
   $k \in [1, K]$ represents the *n_splits* of cross validation; $j \in [1, K] \& j \neq k$
   **for** $k \leftarrow 1$ to $K$ **do**
   $[X_k^a, Y_k^a] = [X_e, Y_e]_j$、$[X_k^t, Y_k^t] = [X_e, Y_e]_k$
   $[Ab_x, Ab_y] =$ get_abnormal($X_k^a, Y_k^a$)
   **end for**
   **return** $[Ab_x, Ab_y]$
6: **function** $[X_{fs}, Y_{fs}] \leftarrow$ Incremental Feature Selection($[X_p, Ab_x], [Y_p, Ab_y]$)
   $[X'_f, Y'_f] \leftarrow$ Feature Selection($[X', Y']$)
7: **function** $[X_f, Y_f] \leftarrow$ ADASYN($X_{fs}, Y_{fs}$)

8: **Output:** Confusion matrix $Q([X_f, Y_f], [X'_f, Y'_f])$

This research uses a two-stage model that combines machine learning and deep learning techniques for practical classifications. Figure 4 illustrates the overall process of the proposed TrailGate framework. The initial classification stage utilizes a Random Forest classifier to identify whether samples are predicted as attacks. This procedure is followed by the secondary classification using a deep learning model. Incorporating Random Forests in the first stage can capture essential patterns in network traffic, which enhances the initial recognition accuracy of abnormal traffic and filters out more representative samples for further analysis in the second stage. Through the two-stage design, the system ensures initial and efficient filtering. This design enhances its capability to detect complex attack behaviors accurately and significantly improves the overall model's performance and applicability in practical applications. The variables are defined in Table 5, and the pseudo-code is detailed in Table 6.

The TrailGate source code is freely available at https://www.healthinformaticslab.org/supp/.

### 3.7.1 Data processing

Section 3.1 outlines the dataset and data preprocessing modules, including converting non-numerical features into numerical ones, feature scaling, and other necessary transformations.

### 3.7.2 Feature selection

In this module, we have developed a feature selection framework for heterogeneous datasets. Our feature selection process consists of several steps, as illustrated in Figure 5. The first step involves calculating the information gain for the training set's discrete and continuous features. We rank all the features based on their information gains and compute a PCC for each pair of features. A pre-defined threshold identifies and filters out highly correlated feature pairs, followed by a redundancy removal process. The selection of thresholds is indicated in the ablation experiment in Section 4.6. Subsequently, we randomly split the training set into training and validation subsets by the ratio of 7:3. Then, the training set is augmented with abnormal samples identified through confident learning to expand the sample space coverage. In Section 4.5, we investigate confident learning to enhance the dataset in the ablation experiment, applying the incremental feature selection (IFS) strategy [40, 41] to identify the top-ranked k features with the best prediction performance. However, the top-ranked k features selected for different classification tasks at different stages may vary. The finally selected features are also indicated in the experimental results in Section 4.6.

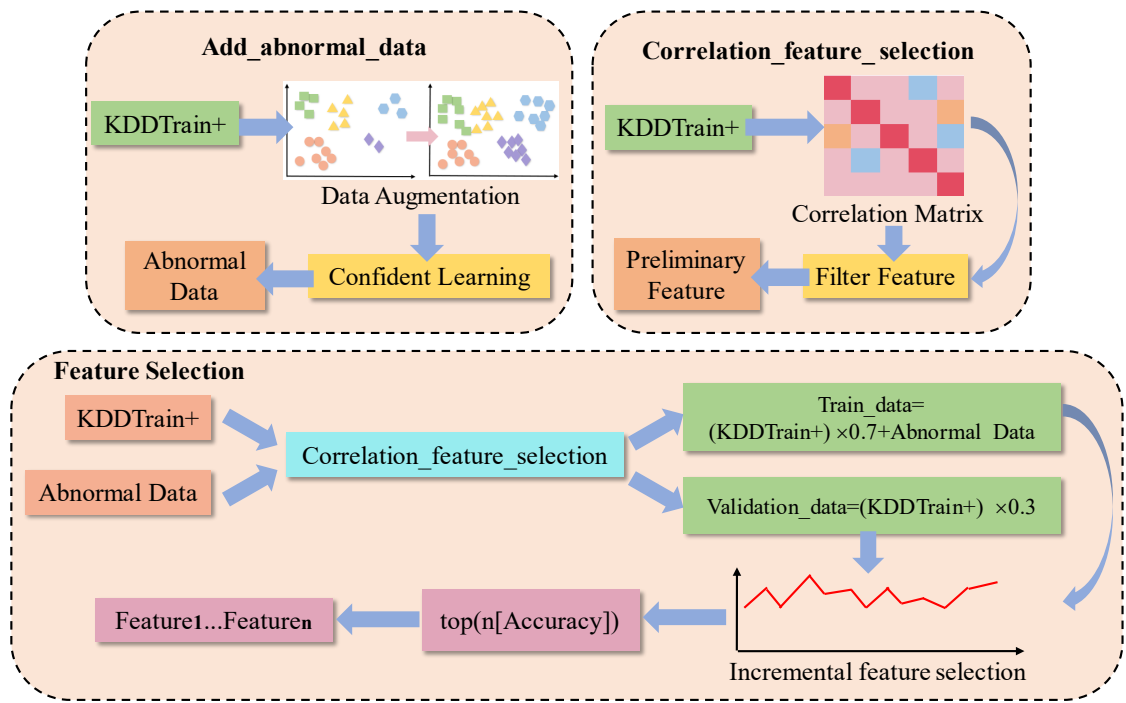

Figure 5. Flowchart of the feature selection process.

### 3.7.3 First-stage classification

In the first classification stage, we use random forests for binary network traffic classification and divide the samples into normal and abnormal categories. This step is robust because it provides a preliminary filtering mechanism for the entire system and makes the subsequent classification process more efficient. In binary classification, the system can focus on a more detailed analysis of different attack types in the second stage and refine the classification model for the attack traffic. This two-stage approach effectively reduces the amount of data for deep learning processing, lowers computational complexity, and improves the detection efficiency of the entire model.

### 3.7.4 Second-stage classification

In the second classification stage, we used a combination model of BiGRU+Transformer Encoder to complete the secondary classification of abnormal traffic. This stage makes BiGRU robust in capturing temporal dependencies in abnormal network traffic and enables the model to gain more decisive insights when analyzing complex time series data. Through BiGRU, the model can utilize temporal information to identify the changing trends of abnormal behavior more accurately.

The Transformer Encoder further optimizes feature representation through its powerful multi-head attention mechanism. It enables the model to focus on the most relevant features and effectively distinguish different attack behaviors. This model supports binary and multi-class classification tasks, differentiating between normal and abnormal traffic, refining classification results, and detecting various attack behaviors (DoS, Probe, U2R, R2L). This flexible architecture enables the model to perform exceptionally in the face of complex and diverse network attacks, responding to various statistical patterns in the network traffic.

The deep learning model undergoes optimization via ten-fold cross-validation to enhance its generalization performance. This optimization aims to correct misclassifications from the first stage, particularly for samples with classification challenges, subsequently integrating the outcomes from both classification stages.

## 3.8 System model

The proposed TrailGate system is a two-stage framework designed to enhance network intrusion detection through feature selection and classification techniques in both stages.

### 3.8.1 First stage: Feature selection and binary classification

Let the input data that has undergone data preprocessing be $X = \{x_1, x_2, \ldots, x_n\}$, where each $x_i$ represents a feature vector with $d$ dimensions. Feature selection is performed to reduce the dimensionality of the input space. We denote the feature selection function as $f_s$, which selects the top-$k$ features based on our proposed general feature selection framework:

$$S_1 = f_s(X) = \{x_1, x_2, \ldots, x_k\}, k \leq d$$

After feature selection, we apply a Random Forest classifier for binary classification. Let $RF(S)$ be the classification function of the Random Forest, where $y_{RF} \in \{0,1\}$ represents the predicted class label (0 for normal, 1 for abnormal samples):

$$y_{RF} = RF(S_1), y_{RF} \in \{0,1\}$$

### 3.8.2 Second stage: Refined feature selection and deep learning classification

The second stage re-selects the features to extract the most critical features for deep learning classification. Let $f_s'$ indicate the second stage of feature selection, which prefers the top-$m$ features:

$$S_2 = f_s'(X) = \{x_1', x_2', \ldots, x_m'\}, m \leq d$$

Then, the refined feature set $S_2$ is passed through a deep learning model composed of BiGRU and a Transformer Encoder for binary and multi-class classification.

For the BiGRU component, the input sequence $S_2 = \{x'_1, x'_2, \ldots, x'_m\}$ is processed in both forward and backward directions. Let $\overrightarrow{h_t}$ and $\overleftarrow{h_t}$ represent the forward and backward hidden states at time step $t$, respectively, computed as follows:

$$\overrightarrow{h_t} = GRU(\overrightarrow{h_{t-1}}, x'_t)$$

$$\overleftarrow{h_t} = GRU(\overleftarrow{h_{t+1}}, x'_t)$$

The final hidden state at time $t$ is the concatenation of the forward and backward states:

$$h_t = [\overrightarrow{h_t}, \overleftarrow{h_t}]$$

Next, the output from the BiGRU is passed through a Transformer encoder, using multi-head attention to capture long-term dependencies in the data. For an input sequence $H = \{h_1, h_2, \ldots, h_T\}$, the attention mechanism is defined as follows:

$$Attention(Q, K, V) = softmax(\frac{QK^T}{\sqrt{d_k}})V$$

where $Q$, $K$, and $V$ are the query, key, and value matrices derived from the input, respectively, and $d_k$ is the keys' dimensionality. This step refines the extracted features and helps in the classification task.

### 3.8.3 Third stage: Final classification

The final output is obtained through fully connected layers. For binary classification, the refined binary label $y_{DL} \in \{0,1\}$ is predicted as follows:

$$y_{DL} = argmax(O_{binary})$$

For multi-class classification, the label $y_{multi} \in \{0,1,2,3,4\}$ is predicted, where the different labels represent different attack types (DoS, Probe, U2R, R2L):

$$y_{multi} = argmax(O_{multi})$$

The system leverages the power of Random Forests for the first stage's binary classification. Then, the results are enhanced with the advanced BiGRU+Transformer model for refined binary and multi-class classifications in the second stage.

The final result is obtained by combining the two-stage classification results to robustly and efficiently classify network traffic into normal and abnormal categories and further distinguish specific attack types.

## 3.9 Evaluation metrics

The NID model's efficacy relies on achieving high accuracy while minimizing false positives. This balance establishes reliable detection performance without overlooking potential intrusion activities. We employ a suite of performance metrics (Table 7) to assess our model's generalization capability. These metrics comprehensively evaluate the model's performance across various aspects. The computation of these metrics relies on the four terms, i.e., *TP* (true positive) denotes the number of correctly classified attack samples; *TN* (true negative) represents the number of correctly classified normal samples, while *FP* (false positive) and *FN* (false negative) refer to the misclassified numbers of normal and attack samples, respectively.

Table 7. Calculation formulas and interpretations for model performance metrics.

| Metrics | Formula | Interpretative Statement |
|---|---|---|
| *Accuracy* | $(TP + TN) / (TP + TN + FP + FN)$ | Proportion of correctly detected intrusions to the total traffic records |
| *Recall* | $TP / (TP + FN)$ | Proportion of correctly detected intrusions to the total actual intrusions |
| *Specificity* | $TN / (FP + TN)$ | Proportion of correctly detected normal records to the total actual normal records |
| *FAR* | $FP / (FP + TN)$ | This term is false alarm rate, defined as the proportion of normal records falsely identified as intrusions to the total actual normal records |
| *Precision* | $TP / (TP + FP)$ | Proportion of actual intrusions correctly identified as such to all detected intrusions |
| *F1-score* | $2 / (1/Precision + 1/Recall)$ | Harmonic mean of Precision and Recall, ranging between 0 and 1 |

# 4 Results and discussion

## 4.1 Performance of TrailGate

The TrailGate framework represents a novel approach to optimize network intrusion detection. By harnessing the power of confidence learning, we can sift through data more effectively, selecting the most relevant features and improving the overall robustness of our model, especially when the inter-class data distributions are inconsistent. The feature selection mechanism reduces the data dimensionality, improves prediction accuracy, and reduces computational costs by allowing deep learning models to focus on relevant patterns. BiGRU processes continuous network traffic data in the second stage by capturing past and future dependencies. At the same time, the Transformer encoder refines the feature representation through its multi-head attention mechanism, allowing the model to prioritize the most critical features. Data augmentation techniques generate synthetic samples to address the data imbalance problem and improve the detection of rare attack types such as U2R and R2L. The two-stage design framework combines feature selection with deep learning models, significantly reducing computational complexity while refining the classification results in the second stage. Finally, rigorous cross-validation affirms the model's robustness, generalization ability, and consistent performance on different datasets and prevents overfitting.

TrailGate is extensively evaluated on independent test sets' binary and multi-class classification tasks. Table 8 signifies the performance metrics of TrailGate in binary classification across two independent test sets. TrailGate demonstrates satisfactory performance on the KDDTest+ dataset across all metrics. It also exhibits a marginal shortfall in Recall on the KDDTest-21 dataset. This discrepancy may stem from a significant distributional divergence between the training and test sets, leading to an increased rate of false positives and the accompanying decline in true negatives.

Figure 6 shows the confusion matrix results of binary classification tasks on two independent test sets, and Figure 7 exhibits the confusion matrix results of multi-class classification tasks on two independent test sets.

Table 9 delineates the results of TrailGate in multi-class classification across the specific attack types on the two test sets. Existing NID methods exhibit marginal weakness in detecting U2R and R2L attacks [23, 24, 42, 43]. Our model performs robustly and achieves high detection accuracy while maintaining low false positives and false negatives rates. However, a slight deficiency is observed in the classification of U2R attacks, impacting the overall model performance. Interestingly, TrailGate demonstrates enhanced proficiency in classifying R2L attacks, surpassing the performance of some existing methods.

The following sections compare TrailGate with the existing NID methods across the two independent test sets.

Table 8. Comparative performance metrics of TrailGate in binary classification across the two independent test sets.

| Metric | KDDTest+ | KDDTest-21 |
|---|---|---|
| Accuracy | 94.10% | 91.59% |
| Recall | 91.71% | 59.67% |
| Specificity | 95.92% | 98.68% |
| FAR | 4.08% | 1.32% |
| Precision | 94.44% | 90.93% |
| F1-score | 93.06% | 72.05% |

Table 9. Comparative performance metrics of TrailGate in multi-class classification across the two independent test sets.

| Dataset | Category | Evaluation matrix | | | | |
|---|---|---|---|---|---|---|
| | | Precision | Recall | Specificity | FAR | F1-score |
| KDDTest+ | Normal | 87.72% | 94.65% | 89.97% | 10.03% | 91.05% |
| | DoS | 93.24% | 86.90% | 96.88% | 3.12% | 89.96% |
| | Probe | 67.61% | 88.27% | 94.91% | 5.09% | 76.57% |
| | U2R | 13.16% | 10.00% | 99.41% | 0.59% | 11.36% |
| | R2L | 80.74% | 52.83% | 98.25% | 1.75% | 63.87% |
| KDDTest-21 | Normal | 56.94% | 78.90% | 86.76% | 13.23% | 66.15% |
| | DoS | 85.98% | 81.32% | 92.33% | 7.67% | 83.58% |
| | Probe | 70.81% | 81.60% | 91.45% | 8.55% | 75.82% |
| | U2R | 10.88% | 10.50% | 98.52% | 1.47% | 10.69% |
| | R2L | 69.00% | 45.10% | 93.87% | 6.13% | 54.55% |

Figure 6. Confusion matrices of the binary classification tasks.

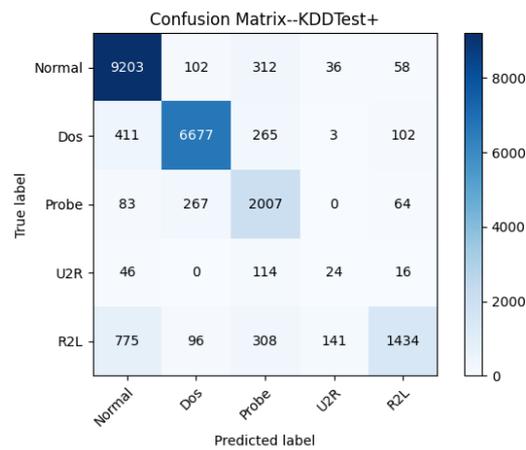

Figure 7. Confusion matrices of multi-class classification tasks.

## 4.2 Comparative analysis of binary classification

Table 10. Comparison of binary classification performance accuracy between TrailGate and existing methods in various test sets. Some studies were not evaluated on all the test sets, and their results are denoted by "-" for the test set where they were not evaluated.

| Method | KDDTest+ | KDDTest-21 |
|---|---|---|
| TDTC [44] | 84.86% | - |
| BAT-MC [45] | 84.25% | - |
| Voting Ensemble [46] | 87.37% | 73.57% |
| Sigmoid_PIO [47] | 86.90% | - |
| Cosine_PIO [18] | 88.30% | - |
| DDQN [48] | 89.78% | - |

| Method | KDDTest+ | KDDTest-21 |
|---|---|---|
| RANet-A [41] | 83.23% | - |
| MFFSEM [49] | 84.33% | - |
| SAAE-DNN [26] | 87.74% | 82.97% |
| SOM-DAGMM [50] | 88.71% | - |
| SigmoidMFO [18] | 88.0% | - |
| CossimMFO [18] | 89.7% | - |
| AE+LSTM [19] | 89% | - |
| Bagging_GBM [20] | 91.57% | 82.35% |
| FACVO-DNFN [21] | 92.86% | - |
| MANormaly [22] | 92.74% | 88.11% |
| SPIP [51] | 81.7% | - |
| **TrailGate (ours)** | **94.10%** | **91.59%** |

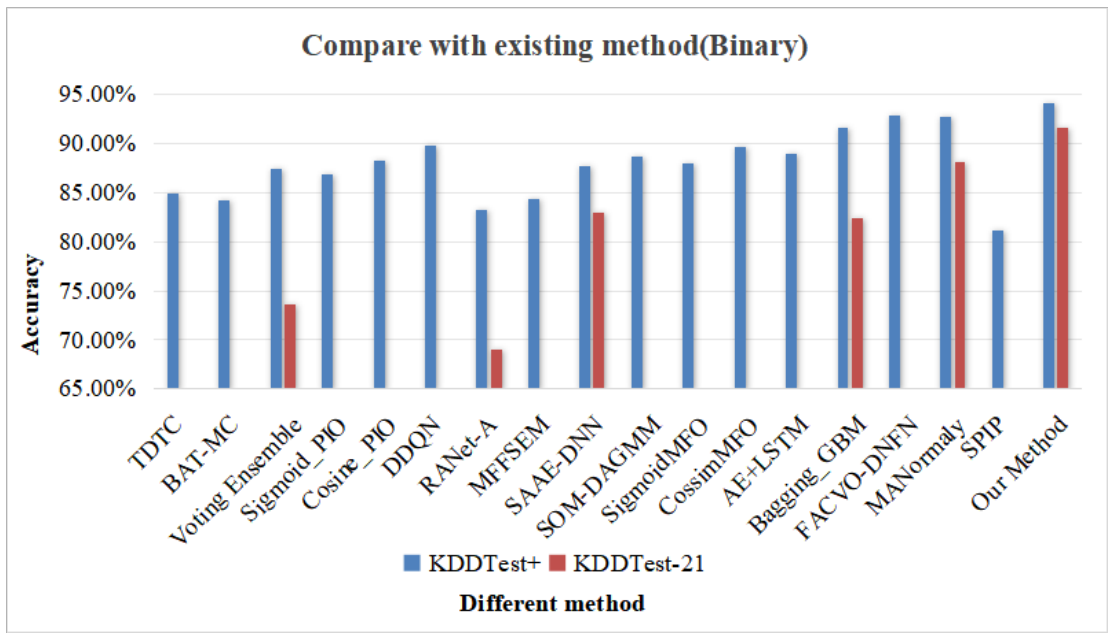

Figure 8. Comparison of binary classification tasks with existing methods

We have thoroughly compared the binary classification performances of TrailGate and several recently published methods, as shown in Table 10. Moreover, a more intuitive comparison of visualization results is shown in Figure 8. TrailGate achieves an impressive accuracy of 94.10% on the KDDTest+ dataset and 91.59% on the KDDTest-21 dataset. These encouraging results demonstrate the model's robust performance on both sets and highlight a marked enhancement, especially in the KDDTest-21 set.

## 4.3 Comparative analysis of multi-class classification

TrailGate is evaluated in multi-class classification on the two independent test sets (Table 11). A more intuitive comparison of visualization results is shown in Figure 9. TrailGate achieves an impressive accuracy of 85.54% on the KDDTest+ dataset and 71.32% on the KDDTest-21 dataset. These results underscore the model's proficiency in handling complex classification situations. As a result, the comparative results of both binary and multi-class classification tasks collectively validate the efficacy of the proposed two-stage model TrailGate across two independent test sets.

Table 11. Comparison of multi-class classification performance accuracy between TrailGate and existing methods across various test sets. Some studies were not evaluated on all the test sets, and their results are denoted by "-" on the test set where they were not evaluated.

| Method | KDDTest+ | KDDTest-21 |
|---|---|---|
| RNN [52] | 81.29% | 64.67% |
| CNN [53] | 79.48% | 60.71% |
| DNNs [54] | 78.5% | - |
| MDPCA-DBN [25] | 82.08% | 66.18% |
| DNN [43] | 77.03% | - |
| DNN+SHAP [55] | 80.3% | - |
| I-SiamIDS [56] | 80.0% | - |
| SAAE-DNN [26] | 82.1% | - |
| MCNN-DFS [57] | 81.4% | - |
| AE+Triplet , B+OVO [42] | 79.13% | - |
| ROULETTE [24] | 83.91% | - |
| DUEN [23] | 82.6% | - |
| **TrailGate (ours)** | **85.81%** | **71.32%** |

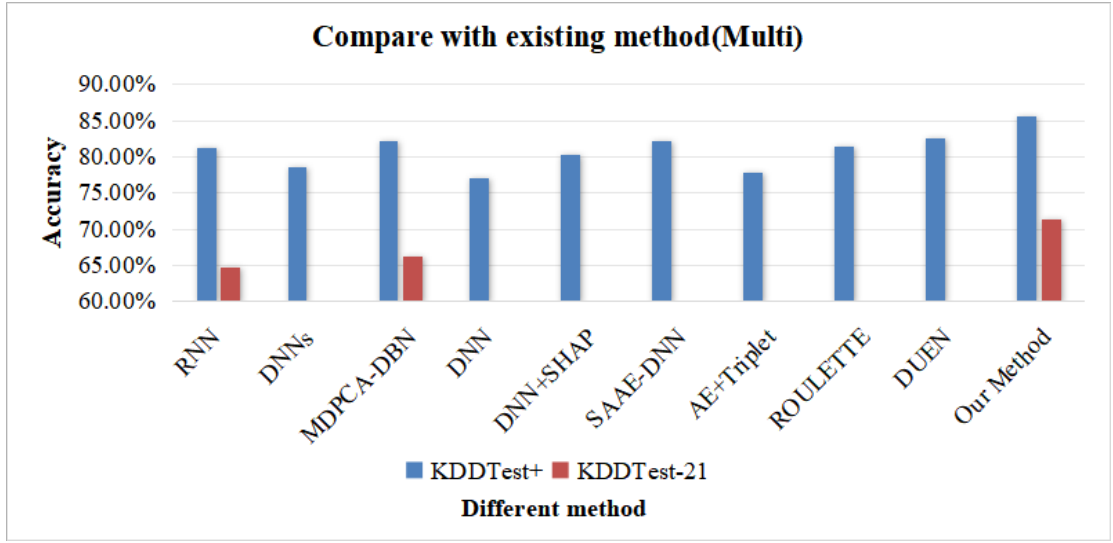

Figure 9. Comparison of multi-class classification tasks with existing methods.

In our further research, we conduct a more in-depth evaluation of the performance of the TrailGate model. Since some related works do not provide detailed classification metrics, we focus on comparing studies that reported F1-scores for the U2R and R2L attack categories. In addition, we compare the overall performance using the Macro F1 score, as shown in Table 12. The results show that when TrailGate detects complex attack types such as U2R and R2L, the detection performance of R2L attacks has been substantially improved, while the performance of U2R attacks is slightly decreased, as shown in Table 13, which many existing models struggle with. Furthermore, TrailGate consistently outperforms other models in terms of overall detection performance, mainly when dealing with imbalanced datasets and underrepresented attack types. These evaluations highlight TrailGate's robustness and capacity to maintain high accuracy across common and rare intrusion types, solidifying its effectiveness in real-world network intrusion detection scenarios.

Table 12. Comparison of multi-class classification performance Macro F1 between TrailGate and existing methods in various test sets. Some studies were not evaluated on all the test sets, and their results are denoted by "-" for the test set where they were not assessed.

| Method | KDDTest+ | KDDTest-21 |
| --- | --- | --- |
| RNN [52] | 63.86% | - |
| CNN [53] | 60.21% | - |
| MDPCA-DBN [25] | 56.96% | 51.76% |
| DNN [43] | 50.8% | - |
| AE+Triplet, B+OVO [42] | 57.50% | - |
| ROULETTE [24] | 61.3% | - |
| DUEN [23] | 64.9% | - |
| **TrailGate (ours)** | **66.61%** | **58.16%** |

Table 13. Comparison of the multi-class classification performance of TrailGate and existing methods on the KDDTest+ test dataset for U2R attack and R2L attack F1-scores. Some studies were not evaluated on all the test sets, and their results are denoted by "-" on the test set where they were not evaluated.

| Method | U2R | R2L |
|---|---|---|
| CNN [53] | 14.44% | 35.45% |
| MDPCA-DBN [25] | 8.38% | 26.51% |
| DNN [43] | 13.00% | 0.00% |
| AE+Triplet, B+OVO [42] | 7.4% | 29.4% |
| ROULETTE [24] | 17.00% | 33.00% |
| DUEN [23] | 26.7% | 42.9% |
| **TrailGate (ours)** | **11.88%** | **64.77%** |

To further evaluate the model's ability to detect emerging threats in realistic scenarios, we constructed a dedicated test set by selecting previously unseen subtypes of attacks from two independent test sets: KDDTest+ and KDDTest-21. These attack types were not present in the training set, making the constructed dataset a suitable proxy for simulating unknown, evolving threats. The resulting test set consists of 7,490 samples and is used to assess the model's performance on unfamiliar attack patterns. Experimental results show that TrailGate achieved an overall prediction accuracy of 70.37% on this emerging threat dataset. The F1-scores for each attack type are illustrated in Figure 10, indicating that TrailGate maintains relatively strong detection performance even for unseen threats. While its performance on U2R attacks was comparatively weaker, it demonstrated robust capability in detecting R2L attacks, achieving an accuracy of 68.37%. These findings provide concrete empirical support for the generalization ability of TrailGate in emerging threat detection, further validating its robustness and practical adaptability in handling continuously evolving cybersecurity threats.

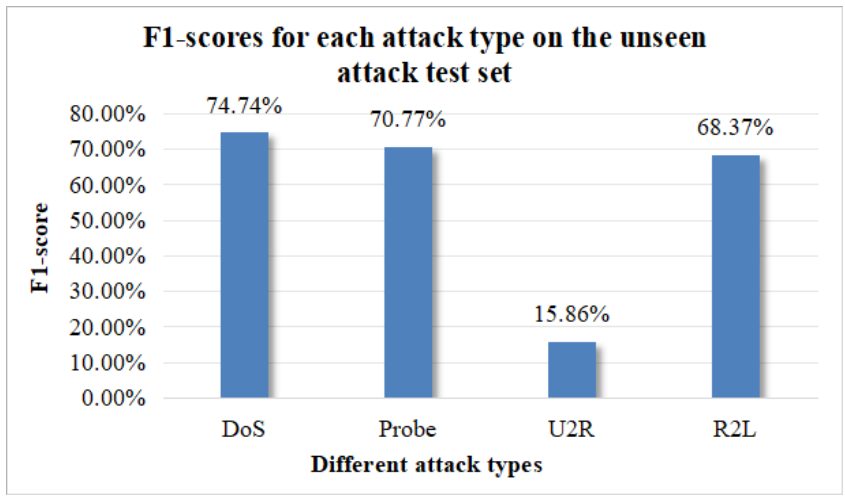

Figure 10. F1-scores for each attack type on the unseen attack test set

## 4.4 Experimental training process

We adopt a two-stage training strategy for the TrailGate framework, each stage optimized for different tasks to improve the overall performance of network intrusion detection. In the first stage, we use the Random Forest classifier to classify the network traffic data into normal and abnormal traffic. The Random Forest classifier trains the input samples by building multiple decision trees as an ensemble learning-based algorithm. Each decision tree makes independent predictions on the sample and finally outputs the classification result through majority voting. We use n_estimators = 300 (i.e., 300 decision trees) for training, the optimal parameter tuned during the experiment. The parameter tuning process is shown in Section 4.6.

After filtering out the abnormal traffic in the first stage, the second stage refines the classification task and uses the BiGRU + Transformer encoder combination model for training. In the first stage, the abnormal traffic classified is passed as input to the second stage for further analysis. In this stage, ten-fold cross-validation is used to verify the model's robustness and generalization ability, using the cross-entropy loss function to measure the accuracy of the classification task. The optimizer Adam (adaptive moment estimation) is selected to adjust the learning rate dynamically to allow the model to converge quickly during training.

We set the number of epochs to 10 and monitor the model's performance on the validation set throughout the training process. After each epoch, the model's accuracy on the validation set is evaluated, and the model with the highest average accuracy across the validation runs is selected and saved. This approach ensures that we capture the best-performing model based on validation accuracy without prematurely stopping the training, and allows the model to fully explore the parameter space over all 10 epochs while minimizing the risk of overfitting. Figures 11 and 12 show the changes in the average accuracy and average loss of each epoch during the training process for the binary and multi-class classification tasks. In addition, by recording the performance metrics such as accuracy, precision, and recall rate of each round, we can adjust the hyperparameters of the model in real time to ensure its optimal performance in different tasks. Figures 13 and 14 show the changes in Acc and Loss for each fold of the ten-fold cross-validation for the binary classification task and the multi-class classification task during the training process. The tuning of other model parameters is shown in Section 4.6.

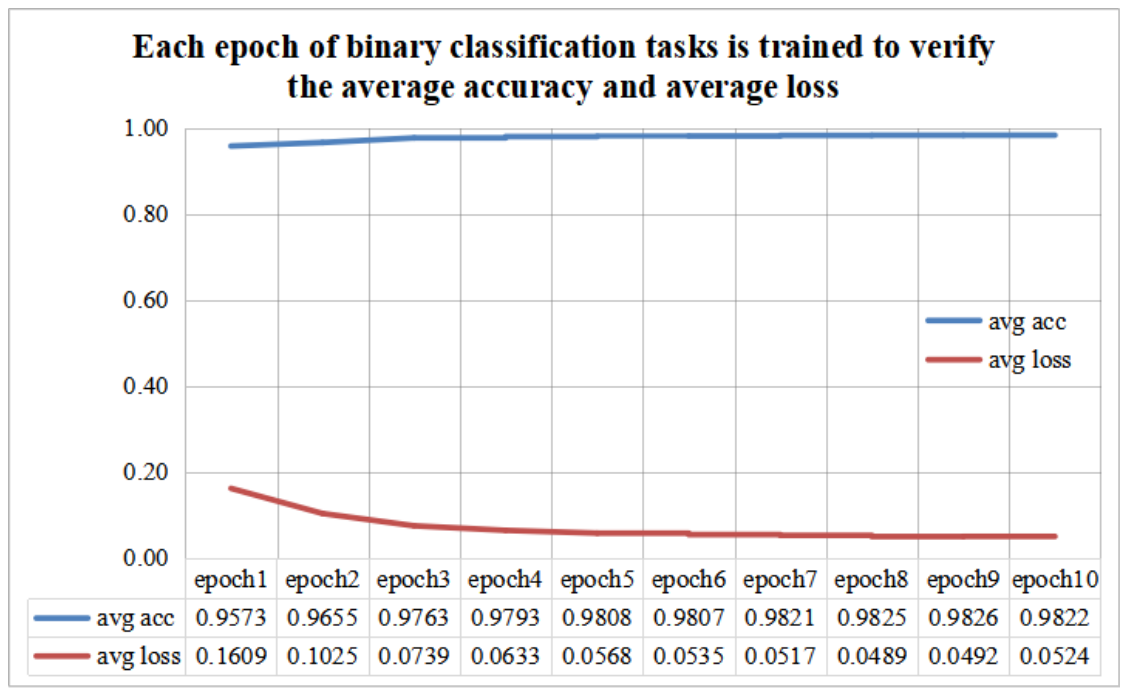

Figure 11. Changes in average accuracy and average loss for each epoch during binary classification task training.

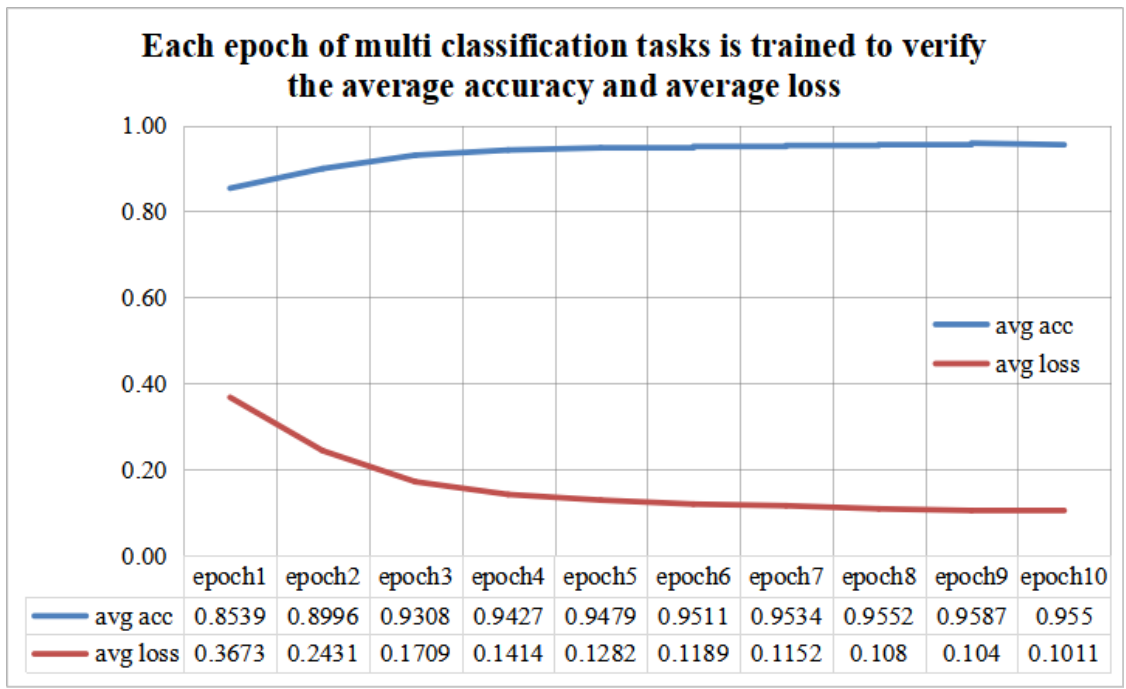

Figure 12. Changes in average accuracy and average loss for each epoch during multi-class classification task training.

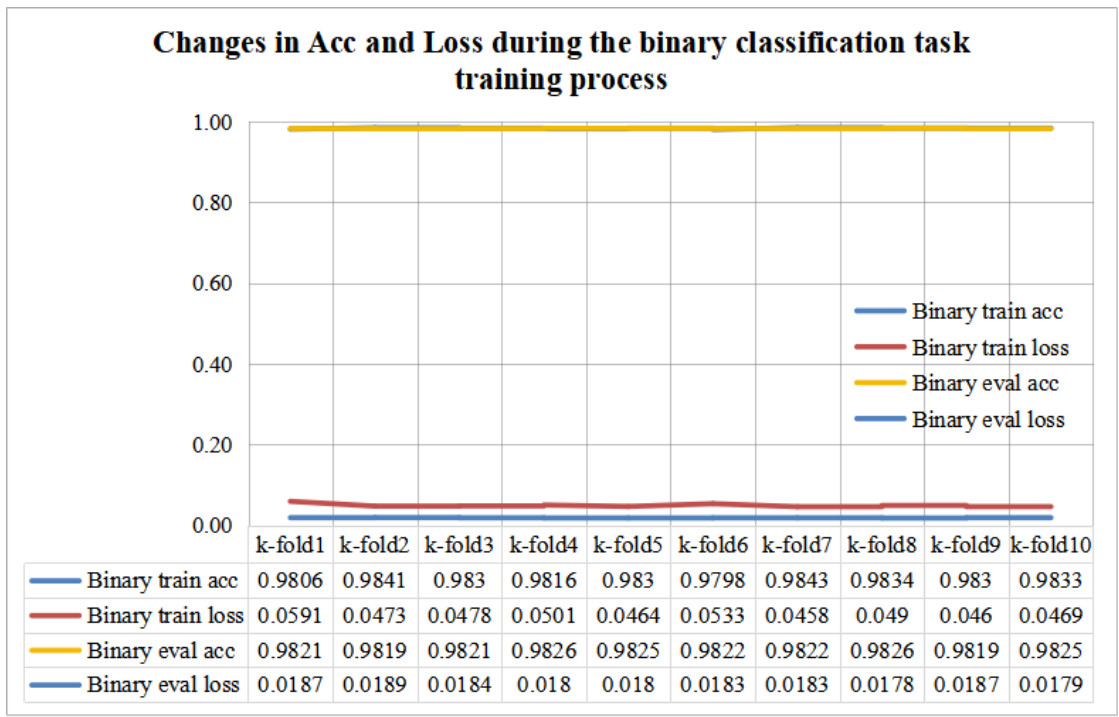

Figure 13. The performance changes in Acc and Loss in binary classification tasks through the k-fold cross-validation strategy.

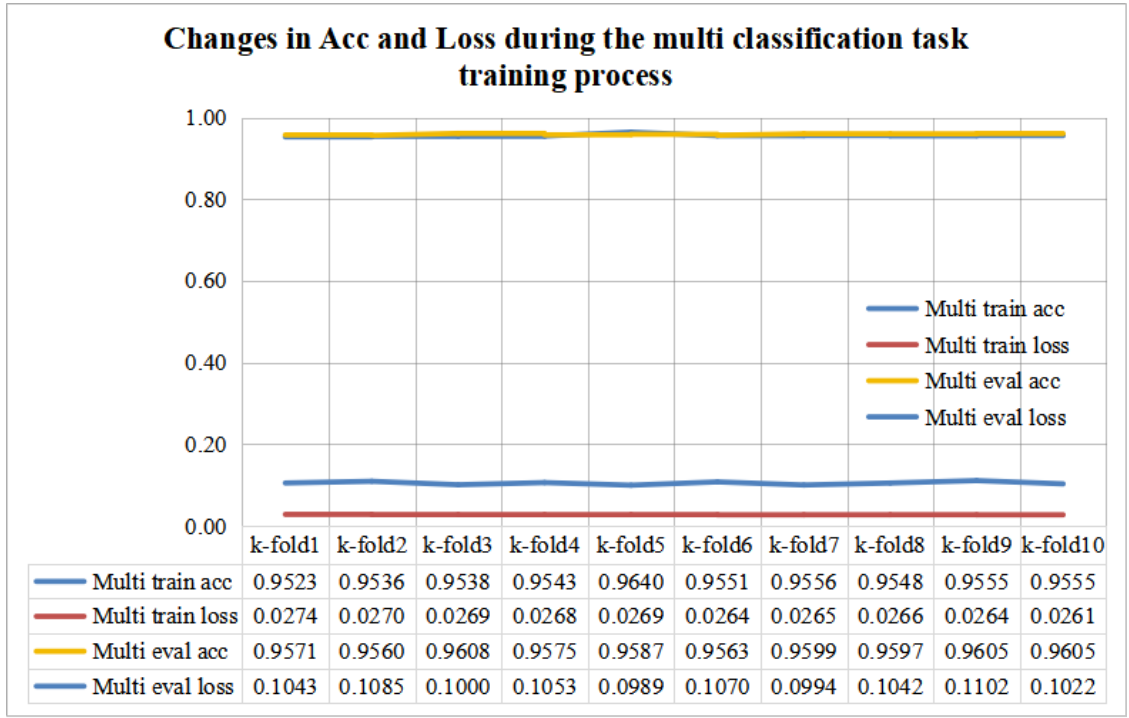

Figure 14. The performance changes in Acc and Loss in multi-class classification tasks through the k-fold cross-validation strategy.

## 4.5 Parameter tuning of the feature selection process

In the feature selection stage, we use the PCCs to filter out highly correlated features and remove redundancy from the dataset. Moreover, we set up ablation experiments for the filtering threshold and employed deep learning models for refining, as shown in Figure 15. The experimental data demonstrates that the binary classification task reaches its peak accuracy at a PCC threshold of 0.7. By contrast, the multi-class classification task exhibits optimal accuracy at a PCC threshold of 0.9. Consequently, we have set the PCC thresholds at 0.7 for binary classification and 0.9 for multi-class classification. We deliberately choose to eliminate a feature if it exceeds this PCC threshold in a feature pair and has a lower information gain than the other feature.

By utilizing the confidence learning strategy, we can effectively pinpoint abnormalities and conduct a series of compelling experiments illustrated in Figure 16 to validate this approach. Our method of selecting abnormal samples is based on the augmented dataset, offering significant advantages in our process. Using sample augmentation, we can expand the sample space effectively and ensure that the newly added samples have similarities with the data in the original training set, thereby improving the model's ability to detect complex and rare attacks. Compared with simply adding the original samples repeatedly, this method is more flexible and can avoid the model's excessive dependence on a few abnormal samples, reducing the risk of overfitting. In addition, sample selection based on augmented data can better address the data imbalance problem, enhance the model's robustness and generalization ability, and improve the overall detection performance. The original training set is randomly divided into the training (T) and validation (V) subsets in a 7:3 ratio. An appropriate number of abnormal samples is added either to the training subset (T+Ab) or to the entire original training set before re-partitioning into new training (T') and validation (V') subsets. Our results indicate that adding abnormal samples to the divided training subset significantly enhances model performance in binary and multi-class classification tasks. This finding suggests that incorporating abnormal samples effectively broadens the sample's space coverage and improves the model's generalization.

Finally, the IFS strategy selects more suitable and generalizable features for different models and classification tasks. We evaluate the accuracy of the validation subset based on the confidence learning partition and determine the optimal number of features through peak accuracy. The IFS strategy performance for binary and multi-class classification tasks is depicted in Figure 17. The binary classification task reaches the best accuracy at the $11^{th}$ feature, and the highest accuracy is achieved at the $12^{th}$ for the multi-class classification task. The final features chosen for the two classification tasks are detailed in Table 14. For the random forest classifier, we calculate the mean accuracy rate of the kth scenario across three consecutive accuracies of the top-ranked ($k$-1), $k$, and ($k$+1) and evaluate the metric difference in the mean accuracy (DMA) between the two scenarios. Each of the first five scenarios exhibits a DMA greater than 0.005, while the sixth scenario does not pass this threshold. We assume that the DMA metric below this threshold indicates minimal information gain from adding this feature, suggesting a negligible contribution to the model. Thus, the Random Forest classifier uses the first five features, such as protocol_type, service, flag, src_bytes, and dst_bytes.

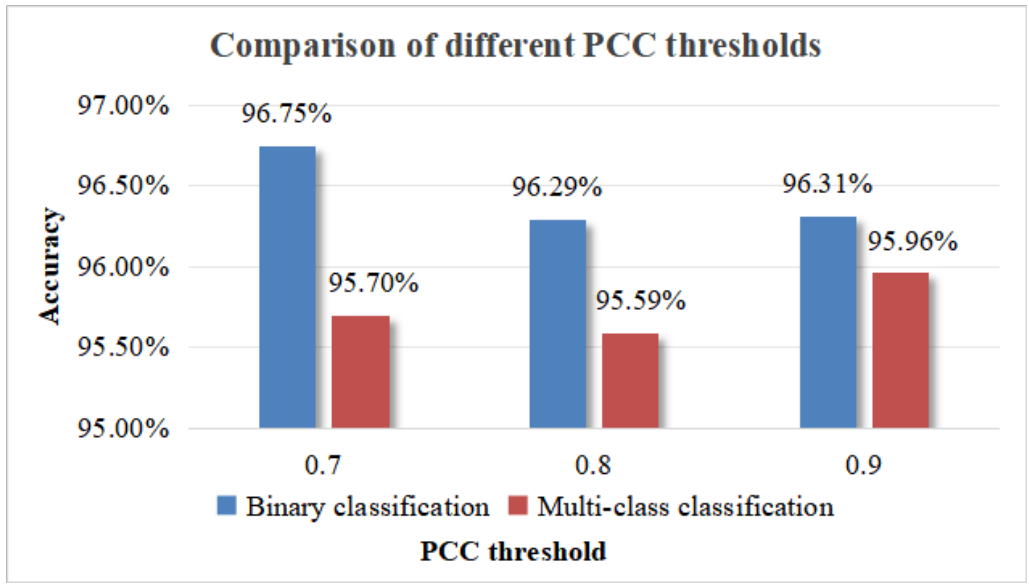

Figure 15. Evaluation of different PCC thresholds for binary and multi-class classifications.

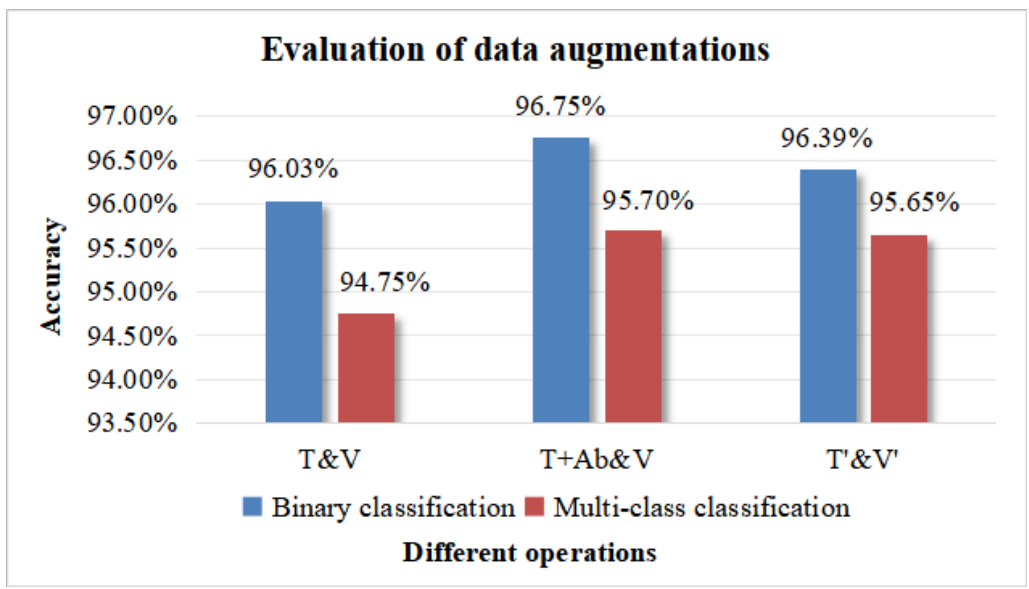

Figure 16. Effectiveness evaluation of augmenting abnormal samples. Ab represents the set of outlier samples selected through confident learning. After dividing the original training set, T and V signify the training and validation subsets, respectively. T' and V' indicate that the new training and validation subsets are split from the training set, respectively, and abnormal samples are added.

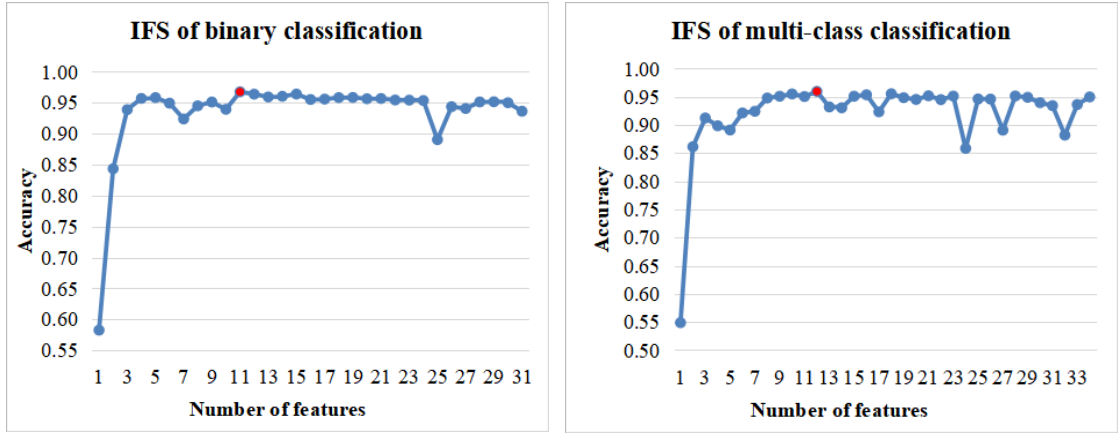

(a)                  (b)

Figure 17. Incremental feature selection (IFS) to find the best number of features. The horizontal axis gives the number of features for each task, and the vertical axis gives the accuracy metric. The data shows the IFS curves for (a) binary and (b) multi-class classification tasks. The peak of each IFS curve is highlighted in red.

Table 14. The final selected features for the two tasks.

| Classifier | Features |
|---|---|
| BiGRU+Transformer (Binary) | protocol_type, service, flag, src_bytes, dst_bytes, same_srv_rate, diff_srv_rate, logged_in, dst_host_srv_serror_rate, dst_host_diff_srv_rate, count |
| BiGRU+Transformer (Multi) | protocol_type, service, flag, src_bytes, dst_bytes, same_srv_rate, diff_srv_rate, logged_in, dst_host_srv_count, dst_host_same_srv_rate, dst_host_srv_serror_rate, dst_host_diff_srv_rate |

## 4.6 Ablation experiment

This section focuses on an ablation experiment to evaluate the significant modules of our proposed two-stage TrailGate. Figure 18 (a) shows that integrating the machine learning model RF in the first stage and the deep learning model BiGRU+Transformer in the second stage yields the most favorable result, with an accuracy of 94.10%. However, the integrated usage of RF in both two stages of TrailGate achieves 85.97% accuracy. The data suggests the essential contribution of BiGRU+Transformer in the second stage of TrailGate.

Figure 18 (b) shows the positive contributions of feature selection and data augmentation training strategies. The binary classification task can be improved by feature selection and data augmentation. Using feature selection (FS) alone provides a 5.95% gain over the original configuration (Raw), and using ADASYN (Aug) alone improves the accuracy by 9.8% over the original configuration (Raw). The combined application of feature selection and data augmentation modules improves by 13.68% over the original configuration, emphasizing the complementary role of balancing data distribution and selecting discriminative features. Figure 18 (b) was completed based on BiGRU+Transformer testing. Therefore, the optimal state of 90.64% in Figure 18 (b) is consistent with the results of using the BT model alone, corresponding to Figure 18 (a).

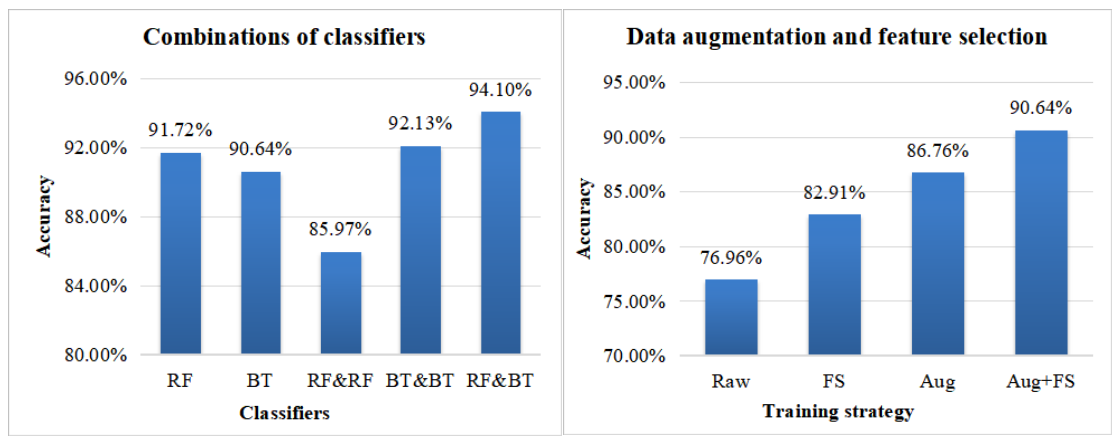

(a)                                                                   (b)

Figure 18. An ablation experiment was based on the binary classification task. The major modules of TrailGate are evaluated, and accuracy is used as the overall performance metric to assess the TrailGate model with different combinations of modules. (a) The histogram of various combinations of classifiers is illustrated. If only the first stage of TrailGate is utilized, RF represents the usage of Random Forest alone, and BT represents the usage of BiGRU+Transformer alone. The combined pairs of two classifiers represent the two classification algorithms used in the first and second stages of the TrailGate model. At the same time, (b) shows the histogram of different training strategies. Raw represents the original KDDTrain+ dataset used to train the prediction model. Aug and FS refer to the employments of the data augmentation and feature selection strategies on the KDDTrain+ dataset, respectively. Aug+FS signifies the combined usage of both Aug and FS strategies.

To more comprehensively demonstrate the impact of data augmentation on the detection performance for each category, we compared the model's performance before and after applying data augmentation in the multi-class classification task. The experiment was conducted on the KDDTest+ dataset using the second-stage deep learning model, and the comparison results are shown in Figure 19. It can be observed that after applying the ADASYN data augmentation method, the F1-scores for all categories improved to varying degrees, with more significant gains observed in originally underrepresented classes. These results further validate the effectiveness of ADASYN in addressing class imbalance and demonstrate that enhancing minority class samples enables the model to learn more discriminative feature representations, thereby improving overall classification performance.

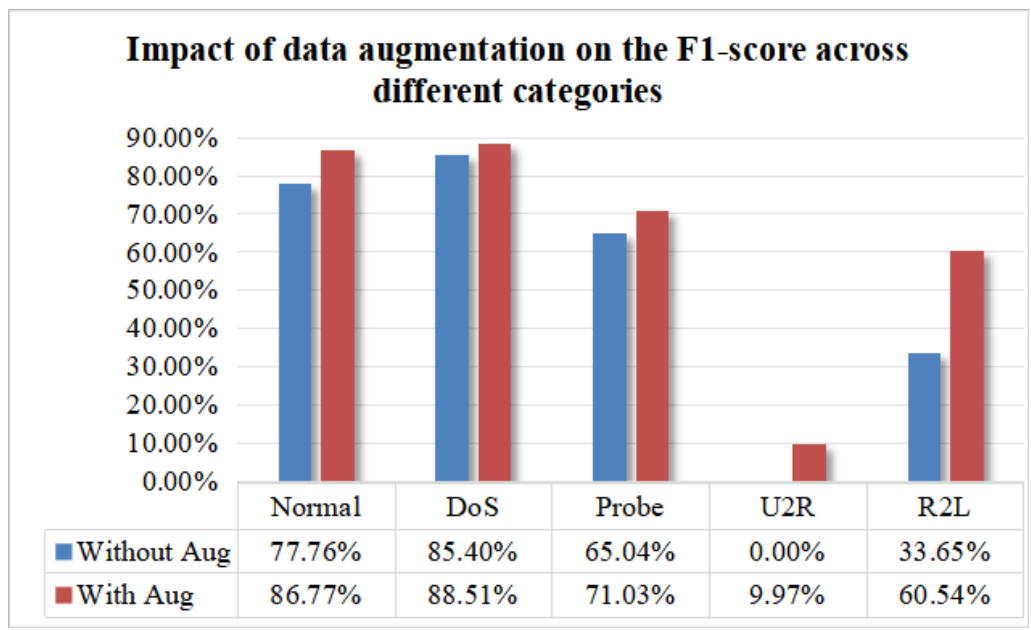

Figure 19. Impact of data augmentation on the F1-score across different categories.

We further compared the impact of different data augmentation ratios on model performance in the multi-class classification task. The experiment was conducted on the KDDTest+ dataset using the second-stage deep learning model. Specifically, we applied data augmentation to all minority classes except the majority class, increasing the number of minority samples to 70%, 80%, 90%, and 100% of the majority class size. As shown in Figure 20, the model achieved the best detection performance when the number of samples in each minority class was increased to match that of the majority class (i.e., 100% augmentation). These results indicate that properly expanding the minority classes can effectively mitigate class imbalance, improve the model's ability to recognize all categories, and enhance its overall generalization and robustness.

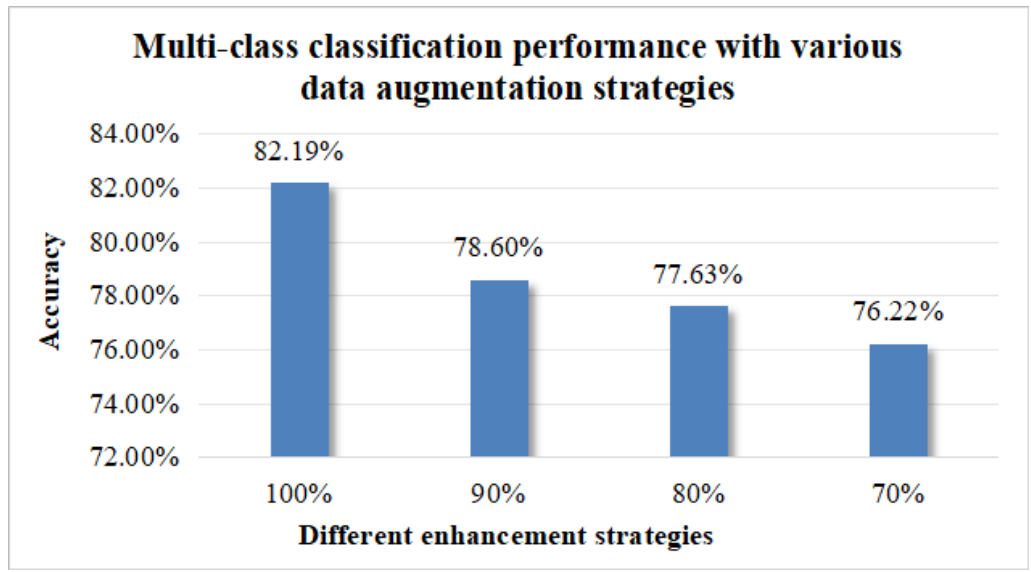

Figure 20. Multi-class classification performance with various data augmentation strategies.

To evaluate the independent contribution of Confident Learning (CL) in the feature selection stage, we conducted a controlled experiment for quantitative assessment. The key aspect of this experiment is to isolate the effect of CL by applying it solely to the feature selection process, rather than during model training, thereby assessing its impact on enhancing model generalization. In the multi-class classification task, the TrailGate model selected 12 features using CL, compared to 21 features selected without CL. As shown in Table 15, the TrailGate model incorporating CL for feature selection achieved superior classification performance on the KDDTest+ test set, outperforming the baseline with 21 features. This performance gain may stem from the inclusion of samples identified as noises during feature selection, which effectively expanded the training set's sample space and mitigated the distribution gap between the training and test sets, thus facilitating the identification of a more discriminative feature subset. These results validate the effectiveness of CL in the feature engineering phase, as it not only improves training efficiency but also enhances generalization by reducing dimensionality.

Table 15. Verify the contribution of CL in the feature selection stage.

| Strategy | Accuracy |
| --- | --- |
| Without CL | 81.03% |
| With CL | **85.81%** |

## 4.7 Parameter evaluation

We tune multiple parameters of the Random Forest and BiGRU+Transformer classifiers to optimize the overall prediction performance of the TrailGate framework. As shown in Table 16, fine-tuning these parameters improves the model's classification accuracy.

In the first stage, we evaluate one of the critical parameters of random forest—the number of base models (n_estimators). Random forest is an ensemble model that uses multiple decision trees as the base models. By experimenting with different n_estimator values (100, 200, 300, 400, 500), we find that as n_estimator increases, the model's accuracy reaches the highest 91.72% when n_estimators = 300 or 500. We choose a more straightforward model form, n_estimators = 300, based on Occam's razor [58-60] to simplify the model while maintaining high accuracy. This decision ensures the model's high performance and reduces its complexity and computational cost, making it more efficient.

In the second stage, we conduct an in-depth evaluation of the batch size (BatchSize) and training epochs (epochs) in the BiGRU+Transformer combined model. Through experiments, we observe that when BatchSize = 512, the model's accuracy reached the highest 98.26% We also find that as the number of training epochs increases, the model's accuracy decreases, indicating that the model may be overfitting after a certain number of training epochs. Therefore, we finally chose the optimal values of BatchSize = 512 and epoch = 9 to balance efficiency and effectiveness. This setting improves the classifier's performance and effectively prevents model overfitting.

Table 16 shows that with the increase in n_estimators, the accuracy of the random forest classifier in the first stage gradually increases and stabilizes after reaching the peak. However, in the second stage, the accuracy of the BiGRU+Transformer classifier increases at different levels.

Our meticulous parameter tuning process, which aims to optimize the performance of the TrailGate framework in binary and multi-class classification tasks, stands as a testament to our commitment. By thoughtfully selecting the values of crucial parameters, we elevate the classifier's accuracy and fine-tune its training efficiency, especially when tackling intricate network intrusion detection challenges.

Table 16. Parameter evaluations of the classifiers in the two stages. The parameter n_estimators represents the number of base models in the ensemble classifier Random Forest. The batch size (BatchSize) and the training epochs (epoch) of the classifier BiGRU+Transformer are evaluated in the second stage.

| | | Random Forest classifier in the first stage | | | | |
|---|---|---|---|---|---|---|
| **n_estimators** | | 100 | 200 | 300 | 400 | 500 |
| **Accuracy** | | 88.35% | 91.71% | 91.72% | 91.71% | 91.72% |
| | | BiGRU+Transformer in the second stage | | | | |
| **BatchSize** | | 32 | 64 | 128 | 256 | 512 |
| **epoch** | 1 | 96.53% | 96.89% | 96.77% | 95.73% | 95.73% |
| | 2 | 97.51% | 97.47% | 97.57% | 96.69% | 96.55% |
| | 3 | 97.81% | 97.71% | 97.82% | 97.63% | 97.63% |
| | 4 | 97.94% | 95.24% | 97.94% | 97.96% | 97.93% |
| | 5 | 97.84% | 95.83% | 98.05% | 98.12% | 98.08% |
| | 6 | 97.89% | 96.32% | 98.06% | 98.16% | 98.07% |
| | 7 | 97.75% | 95.48% | 98.03% | 98.19% | 98.21% |
| | 8 | 96.70% | 95.24% | 97.94% | 98.16% | 98.25% |
| | 9 | 97.61% | 95.24% | 97.98% | 98.22% | **98.26%** |
| | 10 | 97.18% | 95.24% | 98.05% | 98.20% | 98.22% |

## 4.8 Model training time

We evaluate the running time breakdown of the TrailGate's modules on a standard personal computer. Considering the large quantities of network accessing data in the practical situation, the training speed of a NID model is an essential factor for future deployments. We estimate the durations allocated to the significant modules of TrailGate on the binary classification task. Figure 21 shows that the IG and IFS modules consume a substantial portion of the overall duration. This observation highlights an area for potential improvements in future iterations of the TrailGate model. Efficient IG calculation and feature selection are crucial for enhancing the model's practicality and scalability in real-world applications where computational resources and response time are often limited.

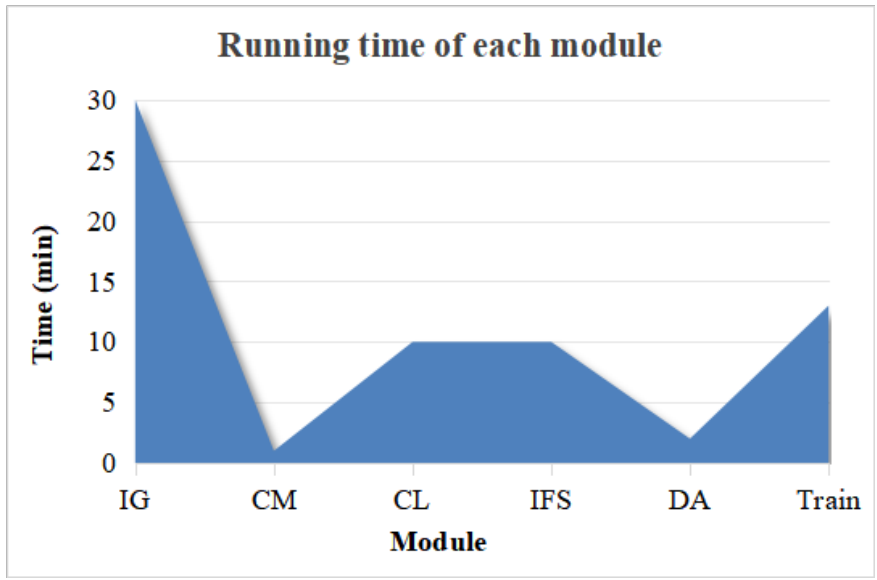

Figure 21. Breakdown of running time of each TrailGate module on the binary classification task. The module IG refers to calculating information gain, and CM refers to calculating the pairwise PCC values. CL is the confident learning module, and IFS refers to incremental feature selection. DA is the data augmentation module. The TrailGate model is trained in the "Train" step.

## 4.9 Comparison with existing techniques

TrailGate offers significant improvements over existing NID techniques in accuracy, efficiency, computational resource optimization, scalability, and robustness, making it a highly effective solution for modern network security challenges. The two-stage approach of the proposed TrailGate combines feature selection and data augmentation methods, achieving significantly improved classification accuracies, especially in detecting rare and complex attack types such as the U2R and R2L attacks. TrailGate was tested on two independent test sets, and binary and multi-class classification tasks were completed with improved accuracy. Compared with the existing work, TrailGate has a higher overall accuracy, achieving an accuracy of 94.10% for binary classification tasks on the KDDTest+ dataset (Table 10), 85.81% for multi-class classification tasks (Table 11), 91.59% for binary classification tasks on the KDDTest-21 dataset (Table 10), and 71.32% for multi-class classification tasks (Table 11). TrailGate reduces the false positive rate by incorporating a confident learning process during the feature selection stage. This process helps identify and select more generalizable features, mitigating the impact of data distribution inconsistencies and leading to performance degradation. By strategically focusing on features resistant to noisy or mislabeled data, the system fortifies its resilience and becomes even more dependable in real-world scenarios. This aspect is crucial as it minimizes false alarms and prevents the wastage of valuable resources.

By employing the feature selection framework, TrailGate effectively screens associated features at both stages, significantly enhancing TrailGate's operational efficiency through reduced input data dimensionality. The feature selection step speeds up the subsequent classification stage. It reduces computational complexity, making it more efficient than deep learning models that run on a complete list of all the original features. By performing lightweight binary classification in the first stage, TrailGate filters out most of the regular traffic in the first stage, allowing the second stage (BiGRU + Transformer) to focus on more challenging cases.

The TrailGate's two-stage approach also facilitates efficient resource utilization, making it feasible to run on personal laptops without server-level hardware. Although the computational cost of the deep learning models (BiGRU+Transformer) is relatively high, the initial Random Forest classification significantly reduces the data the deep learning model needs to process. This reduction in data allows the overall computational cost to remain low. It enables the system to operate efficiently on limited hardware resources, such as those available on personal laptops, compared with more complex models that require high-performance servers from the outset.

TrailGate provides significant improvements over existing methods in terms of classification performance metrics. For example, multi-class classification achieves higher precision and recall values for rare attack types such as U2R and R2L, which are typically more challenging to detect. TrailGate also achieves better F1-scores in these attack types than existing methods. Moreover, TrailGate can handle imbalanced data sets more efficiently. By incorporating data augmentation (ADASYN) into the feature selection process, we can mitigate the bias toward the majority class, leading to enhanced performance on the minority class, which is particularly beneficial for addressing the prevalent challenge of network intrusion detection tasks.

The feature selection process makes the proposed TrailGate framework highly scalable to large-scale networks and more adaptable than traditional deep learning models, which often face computational challenges when dealing with large-scale, high-dimensional datasets. The two-stage design and the feature selection module can be independently updated or replaced with more efficient algorithms. Such a structure setting allows the system to evolve and adapt to new requirements without disrupting the framework. With this flexibility, TrailGate can quickly adapt to growing network environments and increasing data complexity.

TrailGate is better equipped to detect novel and previously unseen attack patterns by leveraging machine learning and deep learning techniques. Integrating BiGRU and Transformer allows the system to capture long-term dependencies in the data and makes it more robust to evolving network threats. The modular framework enables it to be easily adapted to different network environments and traffic patterns. It facilitates TrailGate flexibility compared with single-model solutions that may not generalize well across various scenarios.

We further conducted a comparative analysis between the proposed second-stage deep learning model and several mainstream traditional deep learning approaches in the context of multi-class classification tasks. The evaluation covered not only the classification accuracy of each model on the task, but also computational complexity metrics such as training time and the number of parameters. As illustrated in Figure 22, the BiGRU+Transformer model adopted in TrailGate is compared against other classification models in terms of performance and computational requirements. Classification accuracy is based on the final results from the KDDTest+ test set, while the training time refers to the average per-fold duration in a ten-fold cross-validation setting. The results show that the BiGRU+Transformer model significantly outperforms the other models in accuracy, achieving an improvement of approximately 2.55% over the BiGRU-only variant. This gain highlights the effectiveness of incorporating the Transformer component in capturing long-term dependencies and complex feature representations. Although the BiGRU+Transformer model has the highest parameter count among all evaluated models, its training time remains comparable to the others and within an acceptable range, indicating that the improved performance does not come at the cost of a substantial computational burden. Overall, the BiGRU+Transformer model achieves superior classification performance while maintaining reasonable computational efficiency, demonstrating its practicality and effectiveness in handling complex network traffic data.

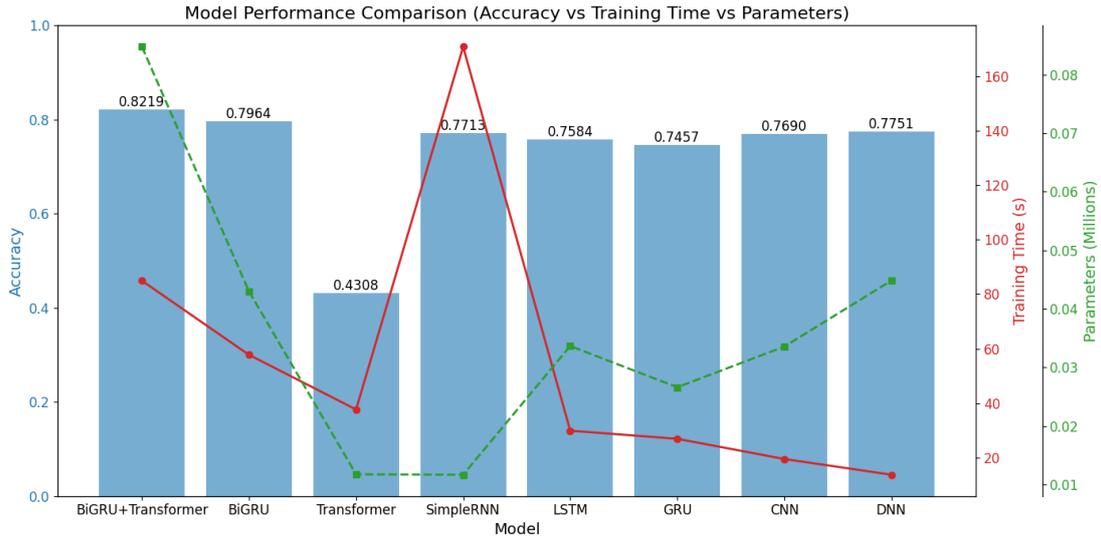

Figure 22. Model performance comparison.

To rigorously evaluate the performance of the proposed TrailGate model and demonstrate the statistical significance of its improvements, we conducted multi-class classification experiments on the KDDTest+ dataset, comparing TrailGate against six baseline models (BiGRU, LSTM, RNN, CNN, GRU, and DNN). Each model was run 10 times with random initialization, using 10-fold cross-validation in each run, and paired statistical tests were performed to assess the significance of the results. Figure 23 presents box plots comparing the test accuracy of all models across the 10 independent runs. In each box plot, the median line represents the average performance, the box shows the interquartile range (IQR), and the whiskers extend to the maximum and minimum values excluding outliers. As shown, TrailGate achieves the highest median accuracy (approximately 0.83), outperforming all baseline models and indicating both strong and consistent performance. The performance gap between TrailGate and other models remains stable across all runs, suggesting the improvements are robust rather than random. To further confirm the statistical significance of these differences, Figure 24 presents a heatmap of *p*-values from pairwise t-tests among the models. All comparisons involving TrailGate yield *p*-values less than 0.001, indicating statistical significance at the 99.9% confidence level. These results verify that the performance improvements brought by TrailGate are not due to chance and highlight its clear advantage over the baseline models.

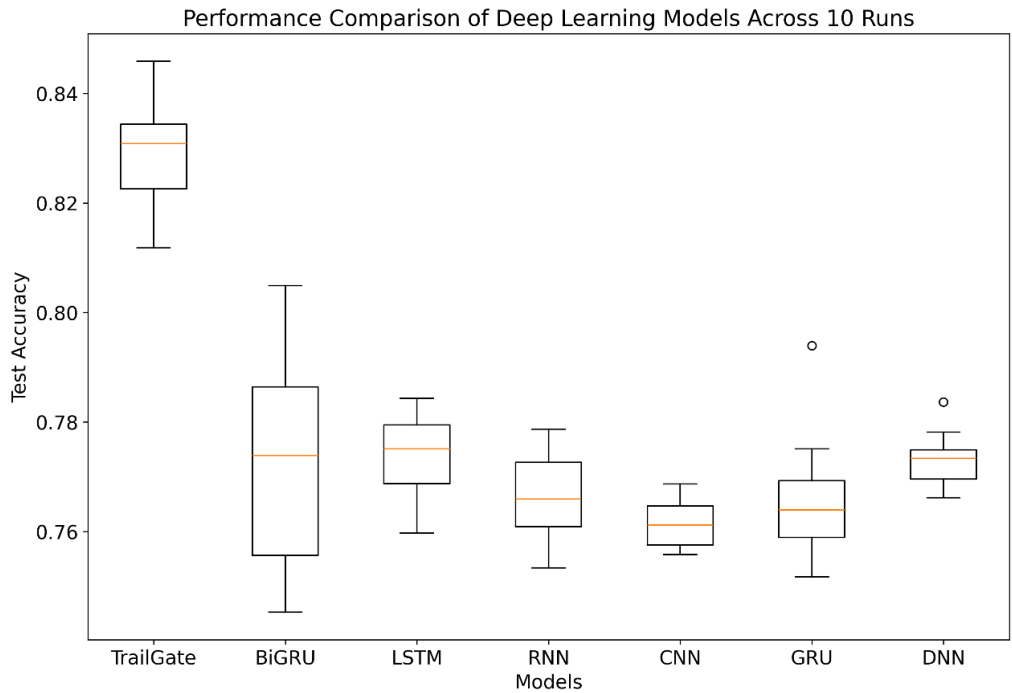

Figure 23. Performance comparison of deep learning models across 10 runs.

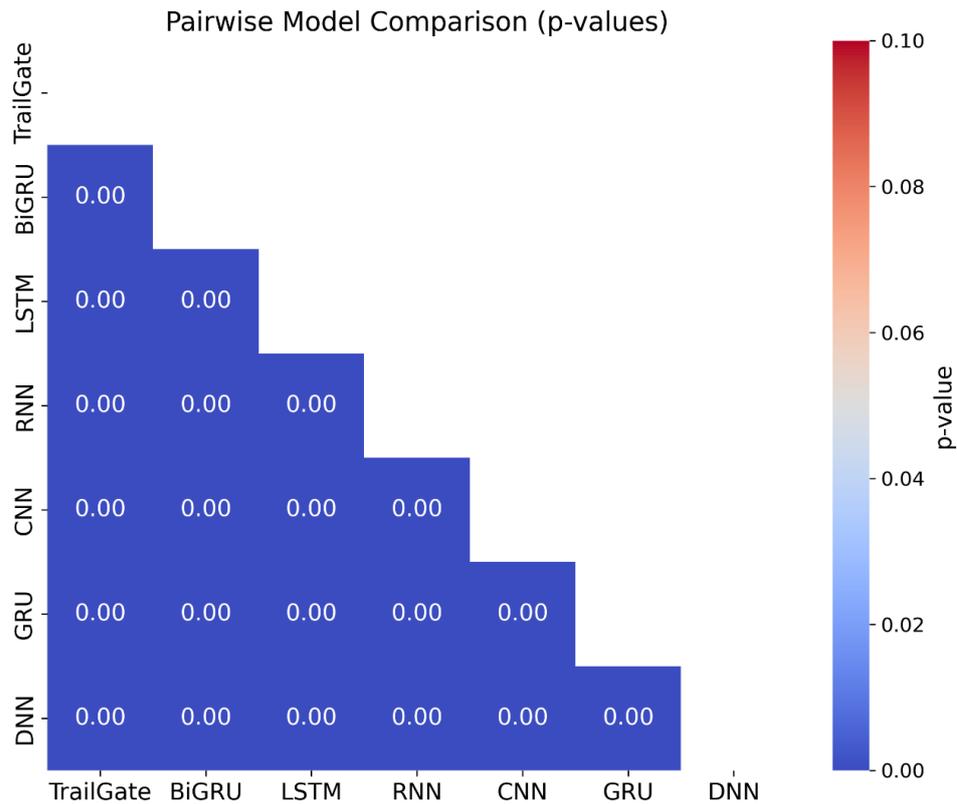

Figure 24. Pairwise model *p*-value comparison.

## 4.10 Generalizability across datasets

To comprehensively evaluate the generalization capability of the proposed model in real-world scenarios, we additionally introduced the UNSW-NB15 dataset as a supplementary benchmark. UNSW-NB15 is a widely recognized modern dataset in the field of cybersecurity, covering a variety of recent attack types and simulating hybrid traffic characteristics that closely reflect real-world network environments. In addition, the dataset offers a standardized and independent train/test split, enabling more objective and fair performance comparisons. The specific data distribution is shown in Table 17. Based on this dataset, we conducted experiments for both binary and multi-class classification tasks and compared the model's performance with that of existing methods, as summarized in Tables 18 and 19, respectively. The results show that the TrailGate model achieved an accuracy of 92.83% in the binary classification task, demonstrating performance that is on par with or even surpasses current mainstream approaches. In the multi-class task, TrailGate also achieved an accuracy of 78.65%, maintaining strong performance. These results provide solid evidence of TrailGate's robustness and adaptability in cross-dataset applications, particularly in its ability to identify high-level discriminative features, and lay a foundation for its practical deployment in broader cybersecurity scenarios.

Table 17. Distribution of UNSW-NB15 dataset.

| Class type | Train dataset | Test dataset |
|---|---|---|
| Normal | 56000 | 37000 |
| DoS | 12264 | 4089 |
| Fuzzers | 18184 | 6062 |
| Analysis | 2000 | 677 |
| Backdoors | 1746 | 583 |
| Exploits | 33393 | 11132 |
| Generic | 40000 | 18871 |
| Reconaissance | 10491 | 3496 |
| Shellcode | 1133 | 378 |
| Worms | 130 | 44 |
| Total | 175341 | 82332 |

Table 18. Performance comparison of TrailGate on the binary classification task using the UNSW_NB15 test dataset.

| Method | UNSW_NB15 test dataset |
|---|---|

| Method | |
|---|---|
| Two-stage ensemble [61] | 91.27% |
| Stacking [62] | 92.45% |
| MFFSEM [49] | 88.85% |
| RF [63] | 83.12 |
| LightGBM [64] | 85.89% |
| EFS-DNN [65] | 88.34% |
| Bagging_GBM [20] | 94.66% |
| SPIP [51] | 86.6% |
| **TrailGate (ours)** | **92.83%** |

Table 19. Performance comparison of TrailGate on the multi-class classification task using the UNSW_NB15 test dataset.

| Method | UNSW_NB15 test dataset |
|---|---|
| MCNN-DFS [57] | 68.5% |
| I-ELM, PCA [66] | 70.7% |
| DNN, Feature selection [67] | 75.6% |
| DNNs [54] | 66.0% |
| Temporal CNN, Attention [68] | 72.9% |
| ROULETTE [24] | 76.4% |
| **TrailGate (ours)** | **78.65%** |

## 4.11 Case study

In this case study, we conducted a more comprehensive analysis building upon existing work. Due to the absence of detailed data in some prior studies, we focused our analysis on the available information. In real-world applications, Denial of Service (DoS) attacks are among the most frequently detected threats. Consequently, we evaluated the F1-score for DoS attacks in the multi-class classification tasks, alongside the F1-score for normal traffic, as illustrated in Table 20. While we previously discussed model performance on rare attack types in Section 4.3, it is important to emphasize here that our model also achieves robust detection performance for more common attack types such as DoS.

The results show that TrailGate consistently maintains high classification accuracy for both normal traffic and prevalent attack types like DoS. This balanced performance is critical, as it ensures the model's efficacy not only in detecting rare, underrepresented attacks but also in accurately identifying more common threats, all while minimizing the False Alarm Rate (FAR). The ability to reduce false positives without compromising accuracy for either frequent or rare attack types enhances TrailGate's reliability in real-world deployment scenarios. As shown in Table 21, we further compared the Macro FAR across multi-class classification tasks, and the results underscored the robustness of our model in reducing false alarms while maintaining strong detection performance across a broad spectrum of attack categories.

Table 20. Comparison of the multi-class classification performance of TrailGate and existing methods on KDDTest+ test dataset for Normal and Dos attack F1-scores. Some studies were not evaluated on all the test sets, and their results are represented by "-" on the test set that they were not evaluated.

| Method | Normal | DoS |
| --- | --- | --- |
| RNN [52] | 83.21% | 89.35% |
| CNN [53] | 81.52% | 87.45% |
| MDPCA-DBN [25] | 82.40% | 81.83% |
| DNN [43] | 81% | 87% |
| AE+Triplet [42] | 86.1% | 88.8% |
| ROULETTE [24] | 83% | 91% |
| DUEN [23] | 86.3% | 90.1% |
| **TrailGate (ours)** | **91.00%** | **91.47%** |

Table 21. Comparison of multi-class classification performance Macro F1 between TrailGate and existing methods in various test sets. Some studies were not evaluated on all the test sets, and their results are represented by "-" on the test set that they were not evaluated.

| Method | Macro FAR |
| --- | --- |
| RNN [52] | 6.22% |
| CNN [53] | 6.75% |
| MDPCA-DBN [25] | 7.00% |
| DUEN [23] | 5.2% |
| **TrailGate (ours)** | **4.06%** |

We further employed t-SNE visualization to analyze the prediction results of the second-stage deep learning model on the KDDTest+ test set. The distributions of correctly and incorrectly predicted samples for both binary and multi-class classification tasks are shown in Figures 25 and 26. The visualization results reveal significant overlaps between certain categories in the feature space, reflecting the inherent complexity of the data distribution. In particular, for misclassified samples, it is common to observe that some samples share highly similar feature representations across different classes, making it difficult for the model to distinguish them accurately. For example, in the multi-class classification task, certain attack types exhibit similar traffic patterns or feature characteristics, making them inherently difficult to separate even in the original feature space, thus increasing the model's classification difficulty. Such feature overlaps and ambiguous boundaries are likely to be prevalent in real-world network environments. Despite the presence of these challenging cases, the majority of samples are still correctly classified, indicating that TrailGate possesses strong discriminative capability and effectively minimizes the false alarm rate while maintaining high detection accuracy. The analysis of these failure cases provides valuable insights for future model improvements. Introducing more discriminative features or designing specialized sub-models to handle ambiguous samples may further enhance the model's performance in complex scenarios and reduce the misclassification rate.

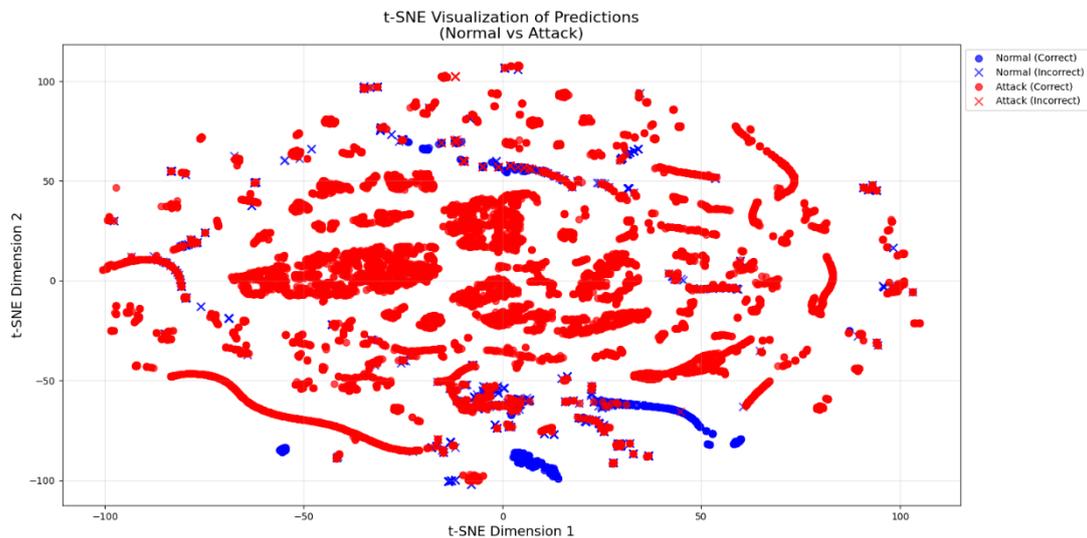

Figure 25. t-SNE visualization of correctly and incorrectly classified samples in binary classification task.

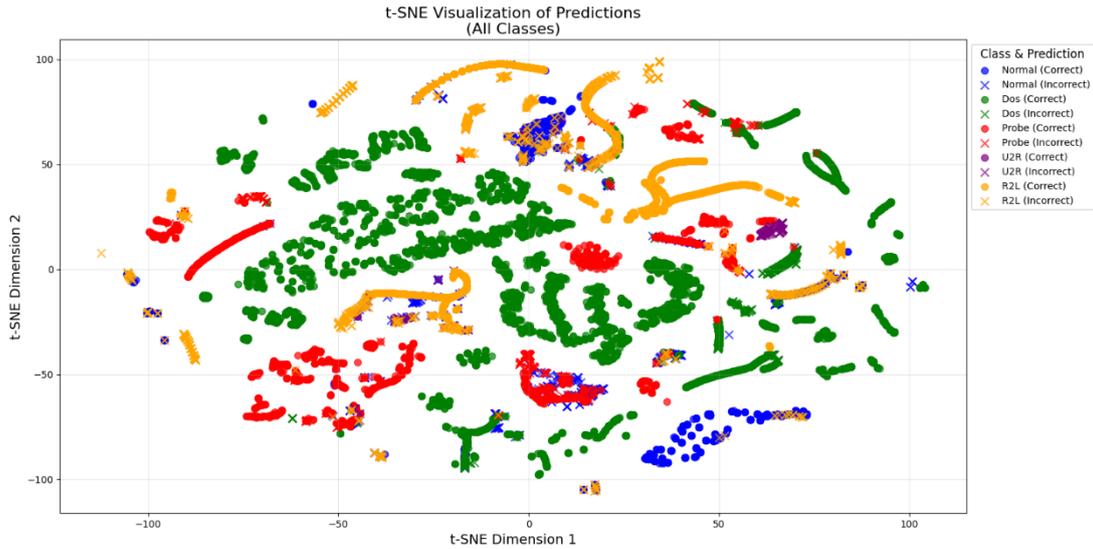

Figure 26. t-SNE visualization of correctly and incorrectly classified samples in multi-class classification task.

## 5 Conclusions

### 5.1 Advantages of the proposed methodology

This study demonstrates that the proposed two-stage framework, TrailGate, offers several advantages in network intrusion detection. By integrating a feature selection module and advanced data augmentation strategies, TrailGate significantly improves detection accuracy, especially for rare and complex attack types such as U2R and R2L. In addition, TrailGate's confident learning screens potential mislabeled data during the feature selection stage and enhances the model's robustness through the reduced impact of noise and imbalanced data.

The implemented feature selection process effectively selects features with excelled prediction performances on both training and independent test sets. This approach mitigates the challenges posed by significant distribution differences between the training and test sets and enhances the model's generalization capabilities. Reducing the number of features positively contributes to enhanced prediction performance and decreased training time.

Moreover, the data augmentation module of our proposed TrailGate framework addresses the data imbalance issue in the training set. Integrating machine learning and deep learning methods also rectifies initial classification errors. This synergy allows for replacing earlier prediction results with more accurate classification results in the second step of TrailGate. Our framework is tested on two independent test sets for both binary and multi-class classification tasks, and the results affirm an improvement in classification performance compared with the existing studies. Furthermore, TrailGate excels at lowering false positive rates while maintaining strong detection performance for both common and rare attack types, which is crucial in real-world scenarios.

TrailGate's versatility enables it to excel in detecting well-known attack types and recognizing novel and evolving threats. Combining BiGRU and Transformer modules allows the framework to capture both short-term and long-term dependencies in network traffic data and enhance its adaptability across diverse network environments. This flexibility makes TrailGate highly effective in intrusion detection scenarios, from common attacks to more sophisticated, previously unseen threats.

## 5.2 Experimental parameters

Several experimental parameters are configured to optimize the performance of the proposed model. The key parameters used in the experiment are as follows: learning rate: 0.001, batch size: 512, iteration times: BiGRU+Transformer model, ten-fold cross-validation, optimizer: Adam optimizer with default parameters, random forest estimator: 300 trees in the set, cross-validation: use ten-fold cross-validation to ensure robustness.

The dataset used for evaluation includes the NSL-KDD dataset, which provides binary and multi-class classification tasks. The KDDTrain+subset is used for training, while KDDTest+ and KDDTest-21 are used as independent test sets.

## 5.3 Limitations

The proposed TrailGate framework has some limitations that must be addressed in future work. One remaining challenge is the computational cost associated with the feature selection process and the two-stage classification model. The feature evaluation and selection steps are critical for enhancing model performance. However, they incur a significant time cost. In addition, using complex deep learning models like BiGRU and Transformer in the second stage demands substantial time and memory resources, making the method less suitable for real-time intrusion detection or environments with limited computational capacity. Another limitation lies in the model's reliance on the quality and representativeness of the training data. The model's performance may degrade when the data is scarce or not reflective of real-world network traffic. Although we have validated our model across multiple datasets, public data in the cybersecurity domain still faces inherent challenges. Most existing benchmark datasets exhibit a generational gap compared to real-world network environments and lack sufficient labeled data for emerging attack vectors. Moreover, many industrial-grade datasets involve privacy and regulatory concerns, often requiring collaborative traffic collection with industry partners. As a result, the availability of high-quality benchmark datasets for academic research remains significantly limited.

## 5.4 Future directions

While the TrailGate framework demonstrates improved capability in network intrusion detection, several fields warrant further exploration to enhance its scalability, efficiency, and applicability. Improving computational efficiency is a crucial focus of future work. By incorporating modern feature engineering techniques, we aim to optimize the feature selection process and reduce time and resource consumption without sacrificing accuracy. In addition, simplifying the deep learning components, such as BiGRU and Transformer, could lower the computational overhead and make the framework more applicable in real-time environments.

Another critical area of exploration is the issue of data imbalance. Intrusion detection systems often struggle with detecting rare or minority attack types, such as U2R and R2L intrusions. To address these issues, future work will investigate the application of generative models for advanced data augmentation and generate synthetic samples to improve detection accuracy for underrepresented intrusion types. This work would help enhance the model's robustness and generalization capabilities, particularly when training data is imbalanced or scarce.

Real-time detection remains a significant challenge. Although TrailGate performs efficiently in batch processing, future research will focus on developing more efficient algorithms and incorporating online learning to enable the framework to process network data in real time. This scenario would involve dynamically updating the model as new data is ingested, making the system more adaptive to emerging threats without compromising detection accuracy.

Another critical direction is expanding the framework's applicability to resource-constrained environments such as IoT devices and edge computing systems. Future iterations of TrailGate could feature a lightweight version that retains high detection performance while optimized for low-power hardware. This area would significantly broaden the framework's utility in many network environments, from high-performance enterprise systems to more resource-constrained devices.

By addressing these issues, TrailGate can develop into a more scalable, efficient, and versatile framework that can meet the needs of modern network environments ranging from high-performance enterprise systems to restricted IoT networks. By continuing to explore new methods such as real-time detection, advanced data augmentation, and adaptive learning techniques, this framework can continue to evolve to meet the growing demands of modern network security environments.

## 5.5 Concluding remarks

The experimental results show that the proposed TrailGate framework significantly improves network intrusion detection methods. TrailGate achieved 94.10% and 91.59% accuracy in binary classification tasks on the KDDTest+dataset and KDDTest-21, respectively, outperforming existing approaches. In the first stage, the feature selection process reduces the dimensionality of the data without affecting accuracy. By contrast, the BiGRU+Transformer model in the second stage is used to refine the classification results further.

TrailGate has also demonstrated strong performance for multi-class classification tasks, especially in detecting U2R and R2L attacks, which other models often overlook. Transformer's attention mechanism enables the model to capture remote dependencies in data, which is crucial in distinguishing similar types of attacks. The proposed method is highly accurate and robust, especially in complex network environments. Future work may focus on optimizing models' computational efficiency, especially in real-time detection scenarios.

## 6 Acknowledgements

This work was supported by the National Natural Science Foundation of China (62072212 and U19A2061), Guizhou Provincial Science and Technology Projects

(ZK2023-297), the Science and Technology Foundation of Health Commission of Guizhou Province (gzwkj2023-565), the Jilin Provincial Key Laboratory of Big Data Intelligent Computing (20180622002JC), the Development Project of Jilin Province of China (No. 20220508125RC), and the Fundamental Research Funds for the Central Universities, JLU. The authors sincerely thank the editor and reviewers for their valuable feedback, which significantly contributed to the improvement of this work during the revision process.